\def\year{2022}\relax
\documentclass[letterpaper]{article} 
\usepackage{aaai22}  
\usepackage{times}  
\usepackage{helvet}  
\usepackage{courier}  
\usepackage[hyphens]{url}  
\usepackage{graphicx} 
\usepackage{natbib}  
\usepackage{caption} 
\DeclareCaptionStyle{ruled}{labelfont=normalfont,labelsep=colon,strut=off} 
\usepackage{algorithm}
\usepackage{algorithmic}

\usepackage{newfloat}
\usepackage{listings}
\floatstyle{ruled}
\newfloat{listing}{tb}{lst}{}
\floatname{listing}{Listing}
\usepackage{amsmath,amsthm,mathtools}
\usepackage{diagbox}
\usepackage{subfigure}
\usepackage{multirow}
\usepackage{url}

\title{Semantic-Aware Representation Blending \\ for Multi-Label Image Recognition with Partial Labels}
\author{
    Tao Pu \textsuperscript{\rm 1}, 
    Tianshui Chen \textsuperscript{\rm 2},
    Hefeng Wu \textsuperscript{\rm 1},
    Liang Lin \textsuperscript{\rm 1}\thanks{Tao Pu and Tianshui Chen contribute equally to this work and share first authorship. Corresponding author is Liang Lin.}
}
\affiliations{
    \textsuperscript{\rm 1} Sun Yat-Sen University, 
    \textsuperscript{\rm 2} Guangdong University of Technology\\
    putao3@mail2.sysu.edu.cn, tianshuichen@gmail.com, wuhefeng@gmail.com, linliang@ieee.org
}

\usepackage{bibentry}

\begin{document}

\maketitle

\begin{abstract}
Training the multi-label image recognition models with partial labels, in which merely some labels are known while others are unknown for each image, is a considerably challenging and practical task. To address this task, current algorithms mainly depend on pre-training classification or similarity models to generate pseudo labels for the unknown labels. However, these algorithms depend on sufficient multi-label annotations to train the models, leading to poor performance especially with low known label proportion. In this work, we propose to blend category-specific representation across different images to transfer information of known labels to complement unknown labels, which can get rid of pre-training models and thus does not depend on sufficient annotations. To this end, we design a unified semantic-aware representation blending (SARB) framework that exploits instance-level and prototype-level semantic representation to complement unknown labels by two complementary modules: 1) an instance-level representation blending (ILRB) module blends the representations of the known labels in an image to the representations of the unknown labels in another image to complement these unknown labels. 2) a prototype-level representation blending (PLRB) module learns more stable representation prototypes for each category and blends the representation of unknown labels with the prototypes of corresponding labels to complement these labels. Extensive experiments on the MS-COCO, Visual Genome, Pascal VOC 2007 datasets show that the proposed SARB framework obtains superior performance over current leading competitors on all known label proportion settings, i.e., with the mAP improvement of 4.6\%, 4.6\%, 2.2\% on these three datasets when the known label proportion is 10\%. Codes are available at \url{https://github.com/HCPLab-SYSU/HCP-MLR-PL}.
\end{abstract}

\section{Introduction}
\begin{figure}[!t] 
  \centering
  \includegraphics[width=0.95\linewidth]{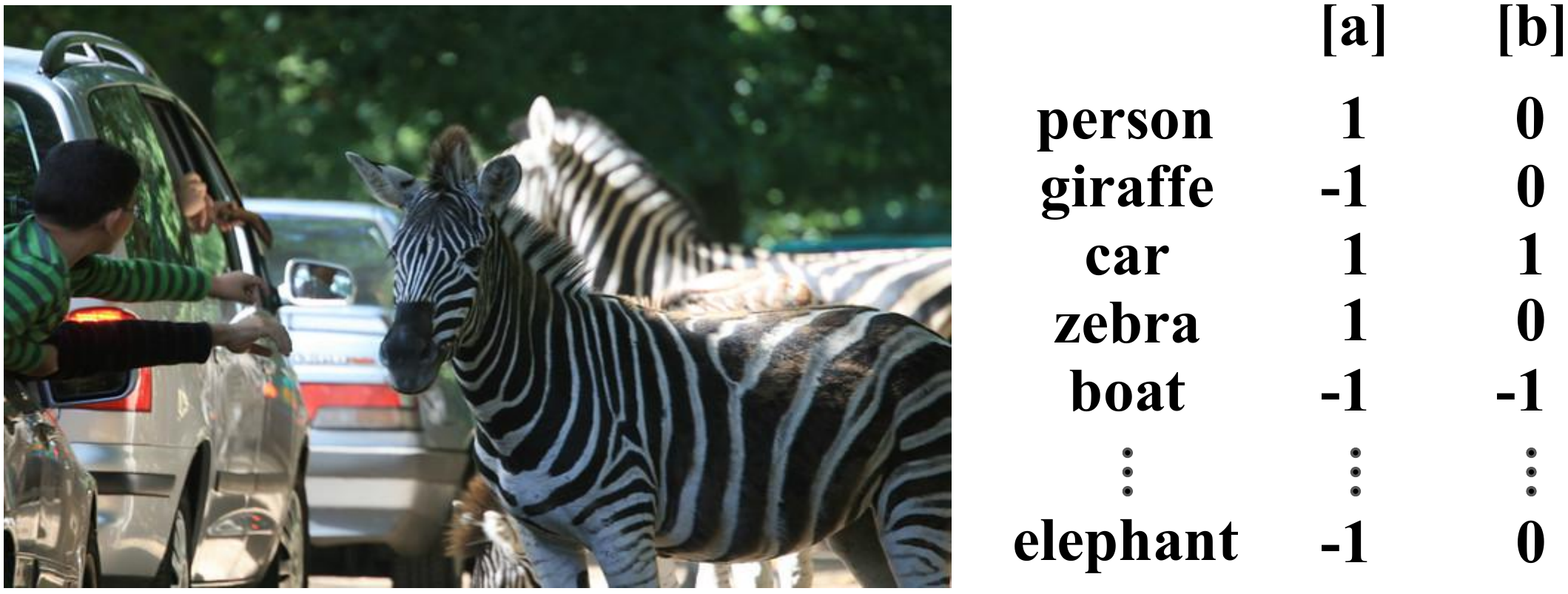}
\caption{An MLR image with complete labels [a], partial labels [b], in which 1 represents the corresponding category exists, -1 represents it does not exist, and 0 represents it is unknown. }     
\label{fig:task-example}     
\end{figure}

Multi-label image recognition (MLR) \cite{Chen2019ML-GCN, Chen2019SSGRL, Wu2020AdaHGNN}, which aims to find out all semantic labels from the input image, is a more challenging and practical task compared with the single-label counterpart. Due to the complexity of the input images and output label spaces, collecting a large-scale dataset with complete multi-label annotation is extremely time-consuming. To deal with this issue, recent works tend to study the task of multi-label image recognition with partial labels (MLR-PL), in which merely a few positive and negative labels are provided whereas other labels are unknown (see Figure \ref{fig:task-example}). MLR-PL is more practical to real-world scenarios because it does not require complete multi-label annotations for each image.

Previous works \cite{Sun2017ICCV, Joulin2016ECCV} simply ignore the unknown labels or treat them as negative, and they adopt traditional MLR algorithms to address this task. However, it may lead to poor performance because it either loses some annotations or even incurs some incorrect labels. More recent works \cite{Durand2019CVPR, Huynh2020CVPR} propose to train classification or similarity models with given labels, and use these models to generate pseudo labels for the unknown labels. Despite achieving impressive progress, these algorithms depend on sufficient multi-label annotation for model training, and they suffer from obvious performance drop if decreasing the known label proportion to a small level.

Fortunately, a specific label $c$ that is unknown in one image $I^n$ may be known in another image $I^m$. We can extract the information of label $c$ from image $I^m$, blend this information to image $I^n$, and in this way complement the unknown label $c$ for image $I^n$. Previous works \cite{Zhang2017Mixup} utilize mixup algorithm to blend two images and generate a new image with semantic information from both images to help regularize training single-label recognition models. However, a multi-label image generally has multiple semantic objects scattering over the whole image, and simply blending two images lead to confusing semantic information. In this work, we design a unified semantic-aware representation blending (SARB) framework that learns and blends category-specific feature representation to complement the unknown labels. This framework does not depend on pre-trained models, and thus it can perform consistently well on all known label proportion settings.

Specifically, we first introduce a category-specific representation learning (CSRL) module \cite{Chen2019SSGRL,Ye2020ADD-GCN} that incorporates category semantics to guide generating category-specific representations. An instance-level representation blending (ILRB) module is designed to blend the representations of the known label $c$ in one image $I^m$ to the representations of the corresponding unknown label $c$ in another image $I^n$. In this way, image $I^n$ can also contain the information of label $c$ and thus this label is complemented. This module can generate diverse blended representations to facilitate the performance but these diverse representations may also lead to unstable training. To solve this problem, a prototype-level representation blending (PLRB) module is further proposed to learn more robust representation prototypes for each category and blend the representation of unknown labels with the prototypes of the corresponding categories. In this way, we can simultaneously generate diverse and stable blended representations to complement the unknown labels and thus facilitate the MLR-PL task. 

The contributions of this work are summarized into three folds: 1) We propose a semantic-aware representation blending (SARB) framework to complement unknown labels. It does not depend on pre-trained models and performs consistently well on all known label proportion settings. 2) We design the instance-level and prototype-level representation blending modules that generate diverse and stable blended feature representation to complement unknown labels. 3) We conduct extensive experiments on several large-scale MLR datasets, including Microsoft COCO \cite{Lin2014COCO}, Visual Genome \cite{Krishna2016VG} and Pascal VOC 2007 \cite{Everingham2010Pascal}, to demonstrate the effectiveness of the proposed framework. We also conduct ablative studies to analyze the actual contribution of each module for profound understanding. 

\begin{figure*}[!t]
   \centering
   \includegraphics[width=0.70\linewidth]{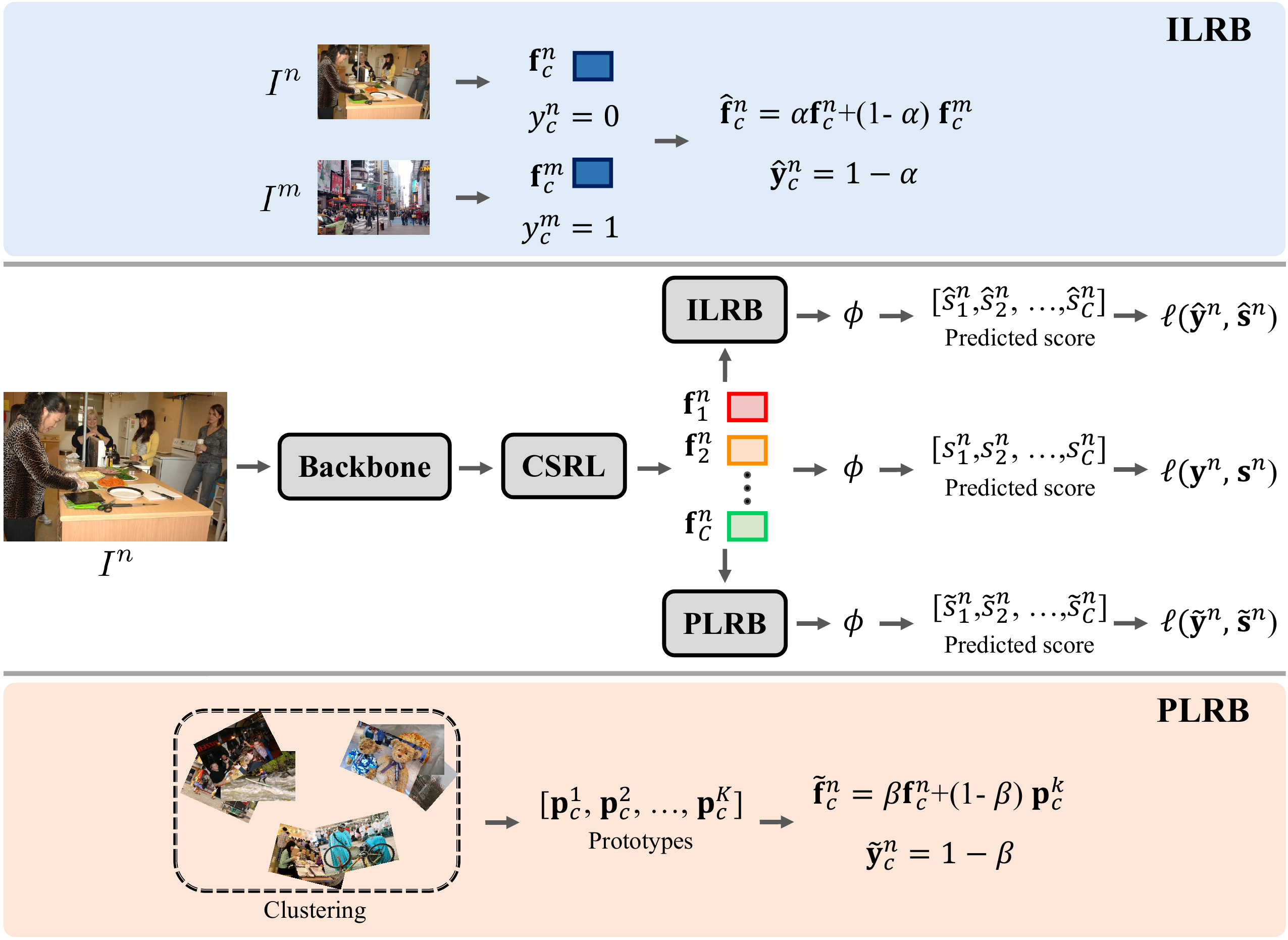}
   \caption{An overall illustration of the proposed semantic-aware representation blending (SARB) framework. It consists of the ILRB and PLRB modules that perform instance-level and prototype-level representation blending to complement unknown labels. The classifier $\phi$ is shared.}
   \label{fig:framework}
\end{figure*}

\section{Related Work}

\noindent{\textbf{MLR with Complete/Partial Labels. }} 
Multi-label image recognition receives increasing attention in the computer vision community due to its wide application to scene recognition \cite{Chen2019RoadScene, zhang2020relational,LiuWL15tcyb}, human attribute recognition \cite{Guo2019Visual, Zhu2017Multi,chen2021cross}, etc. Previous works depend on object localization technology \cite{wei2016hcp} or visual attention mechanism \cite{wang2017multi,chen2018recurrent} to discover discriminative regions and enhance feature representation to facilitate classification. Considering the guidance of semantics to visual representation learning \cite{chen2021hsva}, recent works further introduce category semantics to help learn category-specific discriminative regions \cite{Chen2019SSGRL, Wu2020AdaHGNN}, e.g., Semantic Decoupling (SD) module \cite{Chen2019SSGRL, Wu2020AdaHGNN}, Semantic Attention Module (SAM) \cite{Ye2020ADD-GCN} and Class Activation Maps (CAM) \cite{Gao2021MCAR}. On the other hand, label correlations exist commonly among different categories and these correlations are also important for multi-label recognition. Recent works resort to graph neural networks \cite{AbadalJGLA22csur,ChenCHWLL20aaai} to explicitly model these correlations to learn contextualized feature representation to facilitate multi-label recognition \cite{Chen2019ML-GCN, Chen2019SSGRL, Wu2020AdaHGNN, Ye2020ADD-GCN, Chen2020KGGR}.

Training traditional multi-label image recognition models depends on large-scale datasets with complete annotations per image. To reduce the annotation cost, the current effort \cite{Durand2019CVPR,Huynh2020CVPR} is dedicated to the MLR-PL task, in which merely a few labels are known while the others are known for each image. Earlier works  \cite{Sun2017ICCV, Joulin2016ECCV} formulate MLR as multiple binary classifications, and simply ignore missing labels or treat missing labels negative. Then, they train traditional multi-label models for this task, which leads to poor performance because they lose some data or even incur noisy labels. Inversely, more recent works tend to generate pseudo labels. For example, Durand et al. \cite{Durand2019CVPR} pre-train classification models with the given annotations and generate pseudo labels for the unknown labels based on the trained models. Then, they use both the given and updated labels to re-train the models. Huynh et al. \cite{Huynh2020CVPR} propose to learn image-level similarity models to generate pseudo labels and progressively re-train the model similarly. However, these algorithms rely on sufficient multi-label annotations for model training, leading to poor performance when the known label proportions decrease to a low level.

Different from all these algorithms, our SARB framework learns and blends category-specific feature representation across different images to complement the unknown labels. It gets rid of pre-training models and can obtain consistently well performance on all known label settings.

\noindent{\textbf{Blending Regularization. }}Mixup \cite{Zhang2017Mixup, Yun2019Cutmix, Kim2020Puzzlemix} is recently proposed to blend two input images thus as to generate more diverse samples to regularize training. As a pioneer work, Zhang et al. \cite{Zhang2017Mixup} directly perform pixel-wise blending between two images and it can obtain quite an impressive improvement for single-label image recognition. Cutmix \cite{Yun2019Cutmix} further proposes to randomly cut one region from an image and paste it to another image to generate new samples. Despite achieving impressive performance, these algorithms are very difficult to apply to the multi-label recognition scenarios, because a multi-label image inherently possesses multiple semantic objects scattering over the whole image, and simply blending two images may generate disturbed and confusing information.

Different from the mixup algorithm, the SARB framework proposes to learn and blend category-specific representation, in which the blending is performed between two representation vectors that belong to the same category. In this way, we can utilize the semantic representation of known labels to complement the representation of the unknown labels, and thus to complement these unknown labels.

\section{Semantic-aware Representation Blending}

\subsection{Overview}
In this section, we introduce the proposed SARB framework that consists of two complementary modules that perform instance-level and prototype-level representation blending to complement unknown labels, i.e., the ILRB and PLRB modules. The ILRB module blends the semantic representations of known labels in one image to the presentations of the unknown labels in another image to complement these unknown labels. Meanwhile, the PLRB module learns representation prototypes for each category and blends the representation of the unknown labels of the training image with the corresponding prototypes to complement these unknown labels. Finally, both the ground truth and complemented labels are used to train the multi-label models. Figure \ref{fig:framework} illustrates an overall pipeline of the proposed framework.

Given a training image $I^n$, we utilize a backbone network to extract the global feature maps $\textbf{f}^n$, and then introduce a category-specific representation learning (CSRL) module that incorporates category semantics to generate category-specific representation
\begin{equation}
    [\textbf{f}^n_1, \textbf{f}^n_2, \cdots, \textbf{f}^n_C] = \phi_{csrl}(\textbf{f}^n),
\end{equation}
where $C$ is the category number. There are different algorithms to implement the CSRL module, including semantic decoupling proposed in \cite{Chen2019SSGRL} and semantic attention mechanism proposed in \cite{Ye2020ADD-GCN}. Then we follow previous work \cite{Chen2019SSGRL,chen2019knowledge,chen2018knowledge,chen2021cross} to use a gated neural network and a linear classifier followed by a sigmoid function to compute the probability score vectors
\begin{equation}
  [s^n_1, s^n_2, \cdots, s^n_C]=\phi([\textbf{f}^n_1, \textbf{f}^n_2, \cdots, \textbf{f}^n_C]).
\end{equation}
Based on the learned category-specific semantic representation, the ILRB and PLRB modules are used to complement the feature representation of the unknown labels. We introduce these two modules in the following.

\subsection{Instance-Level Representation Blending}
Intuitively, an unknown label $c$ in image $I^n$ may be known in another image $I^m$. The ILRB module aims to blend the information of label $c$ in image $I^m$ to image $I^n$, and thus image $I^n$ can also have the known label $c$. To achieve this end, we blend the representations that belong to the same category and from different images to transfer the known labels of one image to the unknown labels of the other image.

Formally, given two training images $I^n$ and $I^m$, whose learned semantic representation vectors are $[\textbf{f}^{n}_1, \textbf{f}^{n}_2, \cdots, \textbf{f}^{n}_C]$ and $[\textbf{f}^{m}_1, \textbf{f}^{m}_2, \cdots, \textbf{f}^{m}_C]$, and label vectors are $y^n=\{y^n_1, y^n_2, \cdots, y^n_C\}$ and $y^m=\{y^m_1, y^m_2, \cdots, y^m_C\}$, we blend the semantic representations and labels for each category. For category $c$, the blending process can be formulated as 
\begin{equation}
 \hat{\textbf{f}}^{n}_c=
  \begin{cases}     
   \alpha \textbf{f}^{n}_c + (1-\alpha) \textbf{f}^{m}_c \quad &  y^n_c=0, y^m_c=1, \\
   \textbf{f}^{n}_c \quad & otherwise,
  \end{cases}
\end{equation}

\begin{equation}
 \hat{\textbf{y}}^{n}_c=
  \begin{cases}
   1 - \alpha \quad & y^n_c=0, y^m_c=1, \\
   y^{n}_c \quad & otherwise,\\
  \end{cases}
\end{equation}
where $\alpha$ is the learnable parameter and its initial value is set to 0.5. We repeat the above blending process for all categories, and reformulate them as matrix operations for efficient computing 
\begin{equation}
 {\hat{\textbf{F}}^{n}} = A \textbf{F}^{n} + (1-A) \textbf{F}^{m},
 \end{equation}
\begin{equation}
 {\hat{\textbf{y}}^{n}} = A \textbf{y}^n + (1-A) \textbf{y}^m,
\end{equation}
where $A=[\alpha_1, \alpha_2, \cdots, \alpha_C]$ is a parameter vector; $\textbf{F}^{n}=[\textbf{f}^{n}_1, \textbf{f}^{n}_2,$ $\cdots,$ $\textbf{f}^{n}_C]$ and $\textbf{F}^{m}=[\textbf{f}^{m}_1, \textbf{f}^{m}_2, \cdots, \textbf{f}^{m}_C]$ are the feature matrices for all categories of image $n$ and $m$; $\hat{\textbf{F}}^{n}=[\hat{\textbf{f}}^{n}_1, \hat{\textbf{f}}^{n}_2, \cdots, \hat{\textbf{f}}^{n}_C]$ and $\hat{\textbf{y}}^{n}=[\hat{\textbf{y}}^{n}_1, \hat{\textbf{y}}^{n}_2, \cdots, \hat{\textbf{y}}^{n}_C]$ are the blended semantic representation and label matrix. Then, we use a gated graph neural network and linear classifier followed by sigmoid function to compute the probability score vector $\hat{\textbf{s}}^{n}$.

\subsection{Prototype-Level Representation Blending}
Although the ILRB module can obviously improve the performance, it may disturb the training process because it generates many diverse blended representation for training, especially when the known label proportion is low. To deal with this issue, we further design a PLRB module that learns to generate more stable representation prototypes for each category and blend the representation of unknown labels in image $I^n$ with the prototypes of corresponding categories. 

The prototypes are used to describe the overall representation of the corresponding category. For each category $c$, we first select all the images that have the known label $c$, and then extract the representations of this category, resulting in the feature vectors $[\textbf{f}^{1}_c, \textbf{f}^{2}_c, \cdots, \textbf{f}^{N_c}_c]$. Then, we simply use the K-means algorithm to cluster these feature vectors into $K$ prototypes, i.e., $P_{c}=[\textbf{p}^1_c, \textbf{p}^2_c, ..., \textbf{p}^K_c]$. 

It is expected that the representations of the same category is similar, and thus it can learn more compact distribution to better compute the prototypes for each category. To achieve this end, we utilize contrastive loss for increasing the similarity between $\textbf{f}^n_c$ and $\textbf{f}^m_c$ if images $n$ and $m$ have the same existing category $c$, and decreasing the similarity otherwise. Thus, it can be formulated as
\begin{equation}
\ell^{n,m}_c=
    \begin{cases}
         1-cosine(\textbf{f}^n_c, \textbf{f}^m_c) \quad & y^n_c=1,y^m_c=1,\\
         1+cosine(\textbf{f}^n_c, \textbf{f}^m_c) \quad & otherwise,
    \end{cases}
\end{equation}
where $cosine(\cdot, \cdot)$ represents a function that computes the cosine similarity between the input. The final contrastive loss can be formulated as 
\begin{equation}
\mathcal{L}_{cst}=\sum^{N}_{n=1} \sum^{N}_{m=1} \sum^{C}_{c=1} \ell^{n,m}_c.
\end{equation}

Given an input image $I^n$ whose learned semantic representation vectors $[\textbf{f}^{n}_1, \textbf{f}^{n}_2, \cdots, \textbf{f}^{n}_C]$ and corresponding label vectors $y^n=\{y^n_1, y^n_2, \cdots, y^n_C\}$, we randomly select a label $c$ that is unknown, then randomly select a prototype from $P_{c}$ and blend it with the representation of label $c$, formulated as 
\begin{equation}
 \tilde{\textbf{f}}^{n}_c=
  \begin{cases}
   \beta \textbf{f}^{n}_c + (1-\beta) \textbf{p}^{k}_c \quad & c=random(\{c|y^n_c=0\}) \\
   \textbf{f}^{n}_c \quad & otherwise,\\
  \end{cases}
\end{equation}

\begin{equation}
 \tilde{\textbf{y}}^{n}_c=
  \begin{cases}
   1 - \beta & c=random(\{c|y^n_c=0\})\\
   y^{n}_c  & otherwise,\\
  \end{cases}
\end{equation}
where $\beta$ is a also learnable parameter, and it is initialized as 0.5; $random(\dot)$ represents a random sampling function which means we randomly choose one unknown category to blend semantic representation per image; $k$ is randomly sampled in $[1,...,K]$ and obeys uniform distribution. We repeat the above blending process for all categories, and reformulate them as matrix operations for efficient computing: 
\begin{equation}
 {\tilde{\textbf{F}}^{n}} = B \textbf{F}^{n} + (1-B) \textbf{P}^{k},
 \end{equation}
\begin{equation}
 {\tilde{\textbf{y}}^{n}} = B \textbf{y}^n + (1-B),
\end{equation}
where $B=[\beta_1, \beta_2, \cdots, \beta_C]$ is a parameter vector; $\textbf{F}^{n}=[\textbf{f}^{n}_1, \textbf{f}^{n}_2,$ $\cdots,$ $\textbf{f}^{n}_C]$ and $\textbf{P}^{k}=[\textbf{p}^{k}_1, \textbf{p}^{k}_2, \cdots, \textbf{p}^{k}_C]$ are the feature matrices for all categories of image $n$ and prototype $k$; $\tilde{\textbf{F}}^{n}=[\tilde{\textbf{f}}^{n}_1, \tilde{\textbf{f}}^{n}_2, \cdots, \tilde{\textbf{f}}^{n}_C]$ and $\tilde{\textbf{y}}^{n}=[\tilde{\textbf{y}}^{n}_1, \tilde{\textbf{y}}^{n}_2, \cdots, \tilde{\textbf{y}}^{n}_C]$ are the blended semantic representation and label matrix. Then, we use a gated graph neural network and linear classifier followed by the sigmoid function to compute the probability score vector $\tilde{\textbf{s}}^{n}$.

\subsection{Optimization}

Following previous works, we utilize the partial binary cross entropy loss as the objective function for supervising the network. In particular, given the predicted probability score vector $\textbf{s}^n=\{s^n_1, s^n_2, \cdots s^n_C\}$ and the ground truth of known labels, the objective function can be defined as 
\begin{equation}
\begin{aligned}
\ell(\textbf{y}^n, \textbf{s}^n)=&\frac{1}{\sum_{c=1}^C|y^n_c|}\sum_{c=1}^C[\textbf{1}(y^n_c=1)\log(s^n_c) \\
&+\textbf{1}(y^n_c=-1)\log(1-s^n_c)],
\end{aligned}
\end{equation}
where $\textbf{1}[\cdot]$ is an indicator function whose value is 1 if the argument is positive and is 0 otherwise.

Similarly, we adopt the partial binary cross entropy loss as the objective function for supervising the ILRB module and PLRB module, i.e., $\ell(\hat{\textbf{y}}^{n}, \hat{\textbf{s}}^{n})$ and $\ell(\tilde{\textbf{y}}^{n}, \tilde{\textbf{s}}^{n})$. Therefore, the final classification loss is defined as summing the three losses over all samples, formulated as
\begin{equation}
 \begin{aligned}
  \mathcal{L}_{cls} &= \sum^{N}_{n=1}{ [ \ell(\textbf{y}^n, \textbf{s}^n) + \ell(\hat{\textbf{y}}^{n}, \hat{\textbf{s}}^{n}) + \ell(\tilde{\textbf{y}}^{n}, \tilde{\textbf{s}}^{n}) ] }.
 \end{aligned}
\label{eq:cls-loss}
\end{equation}

Finally, we sum over the classification and contrastive losses of all samples to obtain the final loss, formulated as
\begin{equation}
\mathcal{L}=\mathcal{L}_{cls} + \lambda \mathcal{L}_{cst}.
\label{eq:total-loss}
\end{equation}
Here, $\lambda$ is a balance parameter that ensures the contrastive loss $\mathcal{L}_{cst}$ has a comparable magnitude with the classification loss $\mathcal{L}_{cls}$. Since $\mathcal{L}_{cst}$ is much larger than $\mathcal{L}_{cls}$, we set $\lambda$ to 0.05 in the experiments.

\section{Experiments}

\subsection{Experimental Setting}

\subsubsection{Implementation Details} 
For fair comparison, we follow previous work to adopt the ResNet-101 \cite{He2016ResNet} as the backbone to extract global feature maps. We initialize its parameters with those pre-trained on the ImageNet \cite{Deng2009Imagenet} dataset while initializing the parameters of all newly-added layers randomly. We fix the parameters of the previous 91 layers of ResNet-101, and train the other layers in an end-to-end manner. During training, we use the Adam algorithm \cite{Kingma2015Adam} with a batch size of 16, momentums of 0.999 and 0.9, and a weight decay of $5 \times 10^{-4}$. We set the initial learning rate as $10^{-5}$ and divide it by 10 after every 10 epochs. It is trained with 20 epochs in total. For data augmentation, the input image is resized to 512$\times$512, and we randomly choose a number from \{512, 448, 384, 320, 256\} as the width and height to crop patch. Finally, the cropped patch is further resized to 448$\times$448. Besides, random horizontal flipping is also used. To stabilize the training process, we start to use the ILRB and PLRB modules at epoch 5, and re-compute prototypes of each category for every 5 epochs. During inference, the ILRB and PLRB modules are removed, and the image is resized to 448$\times$448 for evaluation.

\subsubsection{Dataset} We conduct experiments on the MS-COCO \cite{Lin2014COCO}, Visual Genome \cite{Krishna2016VG}, and Pascal VOC 2007 \cite{Everingham2010Pascal} datasets for fair comparison. MS-COCO covers 80 daily-lift categories, which contains 82,801 images as the training set and 40,504 images as the validation set. Pascal VOC 2007 contains 9,963 images from 20 object categories, and we follow previous works to use the trainval set for training and the test set for evaluation. Visual Genome contains 108,249 images from 80,138 categories, and most categories have very few samples. In this work, we select the 200 most frequent categories to obtain a VG-200 subset. Moreover, since there is no train/val split, we randomly select 10,000 images as the test set and the rest 98,249 images are used as the training set. The train/test set will be released for further research.

Since all the datasets have complete labels, we follow the setting of previous works \cite{Durand2019CVPR, Huynh2020CVPR} to randomly drop a certain proportion of positive and negative labels to create partially annotated datasets. In this work, the proportions of dropped labels vary from 90\% to 10\%, resulting in known labels proportion of 10\% to 90\% . 

\subsubsection{Evaluation Metric} For a fair comparison, we adopt the mean average precision (mAP) over all categories for evaluation under different proportions of known labels. And we also compute average mAP over all proportions for a more comprehensive evaluation. Moreover, we follow most previous MLR works \cite{Chen2019SSGRL} to adopt the overall and per-class precision, recall, F1-measure (i.e., OP, OR, OF1, CP, CR, and CF1) for more comprehensive evaluation. We present the formulas of these metrics and detailed results in the supplementary material due to the paper limit. 

\begin{figure*}[!h] 
  \centering    
  \subfigure[]{

  \label{fig:mAP-result-1}
  \includegraphics[width=0.32\linewidth]{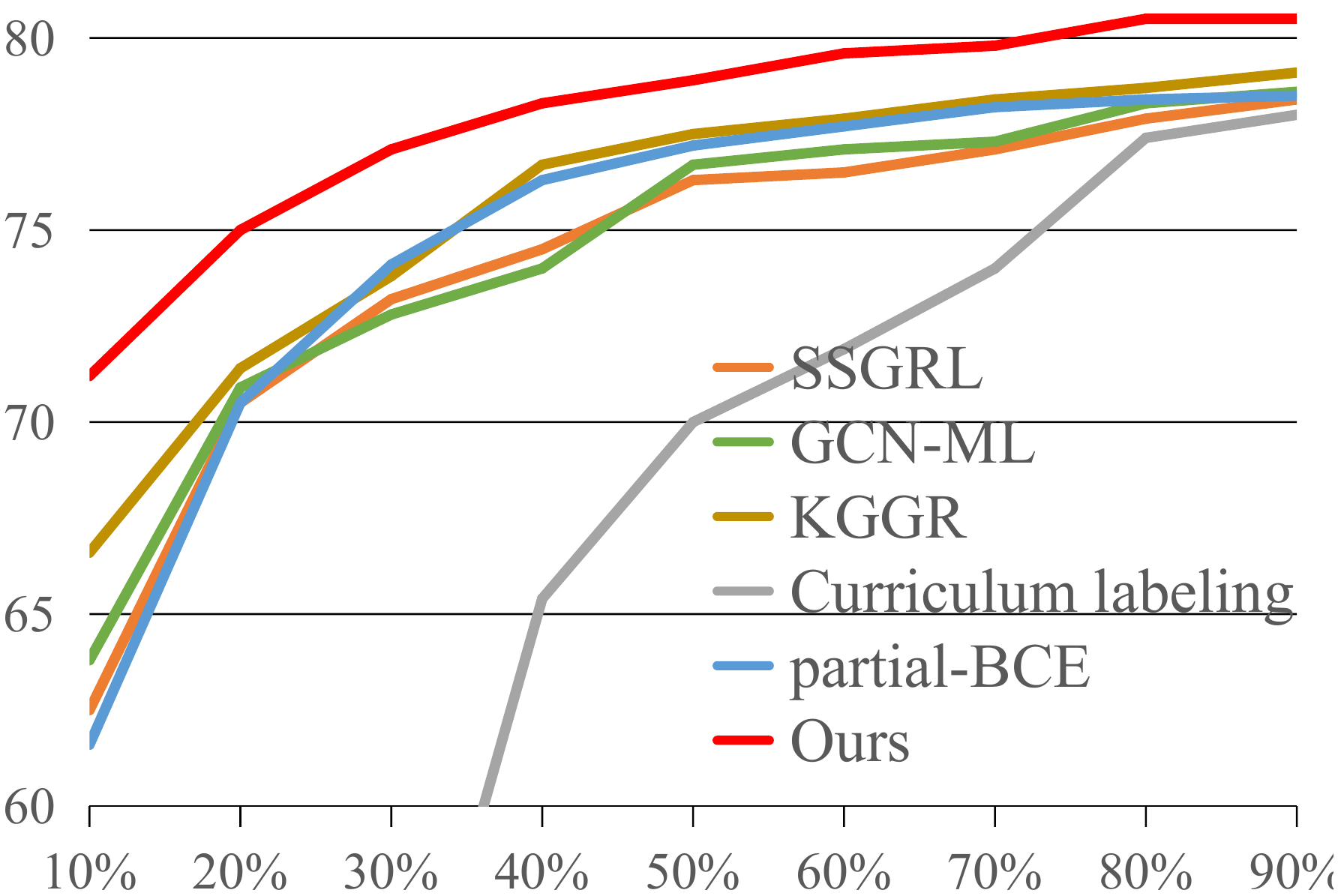}  
  }~     
  \subfigure[]{ 

  \label{fig:mAP-result-2}
  \includegraphics[width=0.32\linewidth]{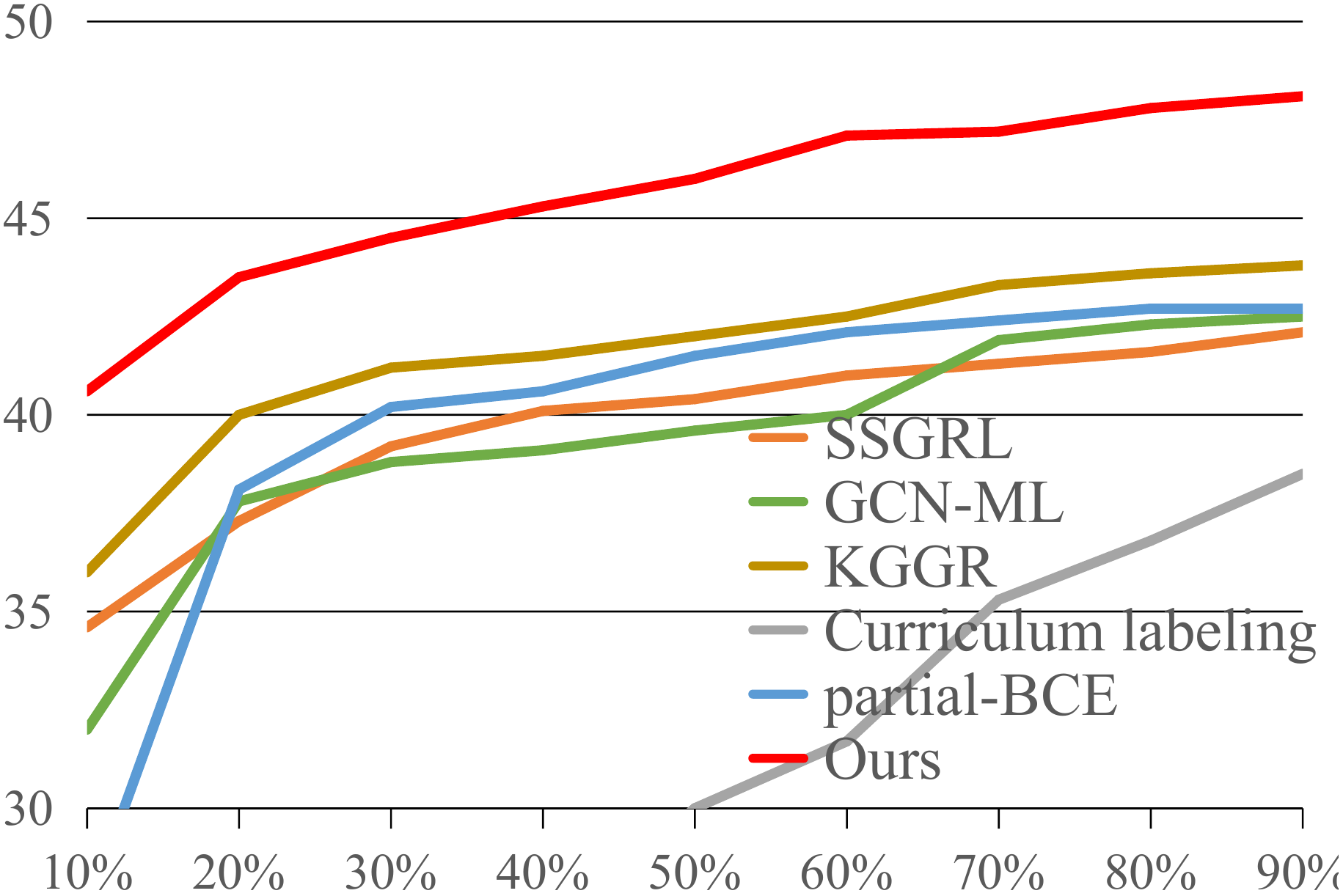}     
  }~    
  \subfigure[] { 
  \label{fig:mAP-result-3}
  \includegraphics[width=0.32\linewidth]{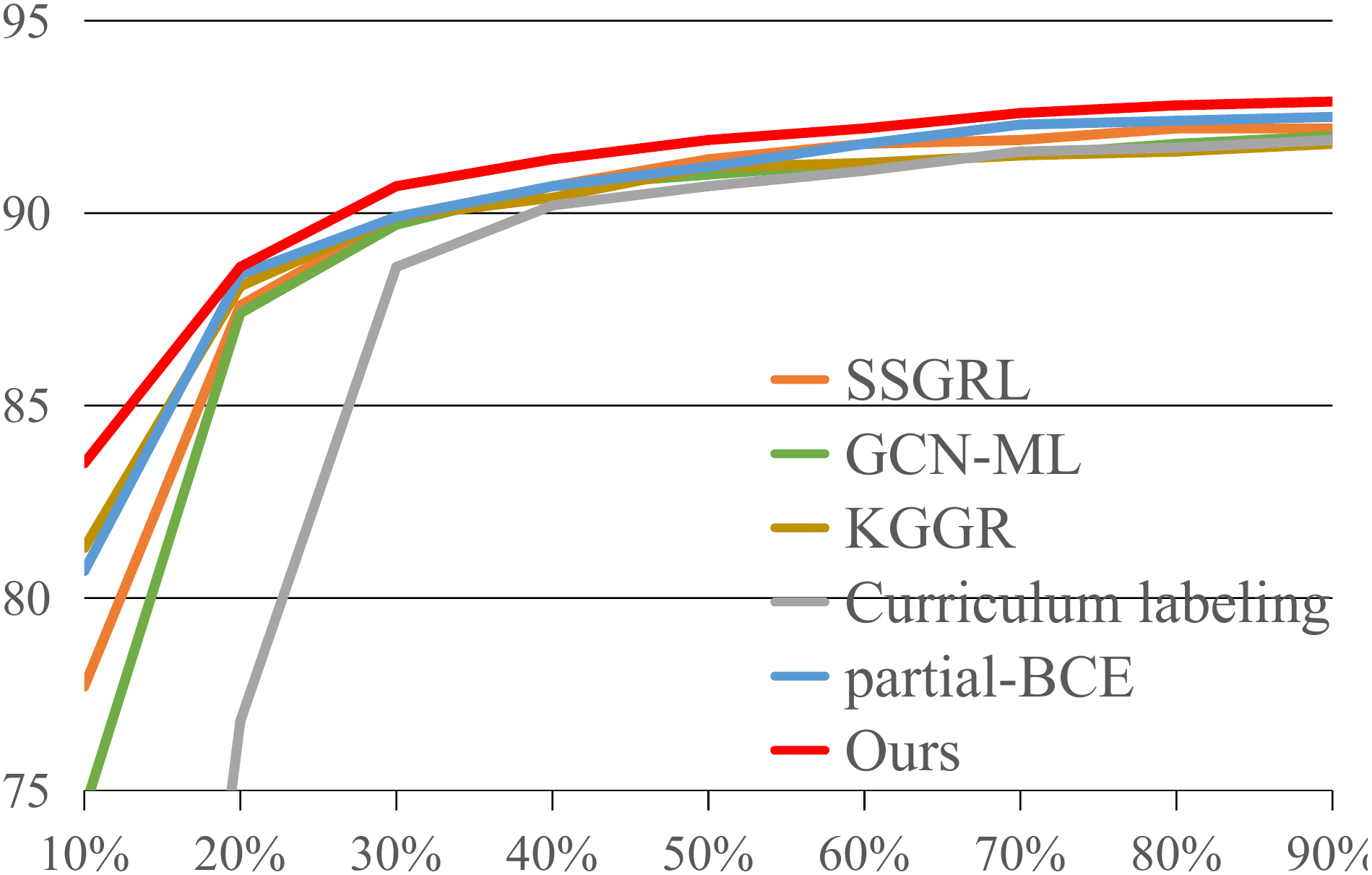}     
  }\vspace{-2ex}
\caption{The mAP of our SARB framework and current state-of-the-art competitors on the settings of known label proportions of 10\% to 90\% on the MS-COCO (left), VG-200 (middle) and Pascal VOC 2007 (right) datasets. Best viewed in color.}
\label{fig:mAP-result}     
\end{figure*}

\begin{table*}[!h]
  \centering
  \begin{tabular}{@{\hspace{0.5ex}}c@{\hspace{0.5ex}}|@{\hspace{0.5ex}}c@{\hspace{0.5ex}}|ccccccc}
  \hline
  \centering Datasets & Methods & Avg. mAP & Avg. OP & Avg. OR & Avg. OF1 & Avg. CP & Avg. CR & Avg. CF1 \\
  \hline
  \hline
  \centering \multirow{6}*{MS-COCO} & SSGRL & 74.1 & 86.3 & 64.8 & 73.9 & 82.1 & 58.4 & 68.1 \\
  \centering ~ & GCN-ML & 74.4 & 85.2 & 64.2 & 73.1 & 81.8 & 58.9 & 68.4 \\
  \centering ~ & KGGR & 75.6 & 84.0 & 65.6 & 73.7 & 81.4 & 60.9 & 69.7 \\
  \centering ~ & Curriculum labeling & 60.7 & 87.8 & 51.0 & 61.9 & 60.9 & 40.4 & 48.3 \\
  \centering ~ & partial-BCE & 74.7 & \textbf{86.7} & 64.7 & 74.0 & \textbf{83.1} & 58.9 & 68.8 \\
  \centering ~ & Ours & \textbf{77.9} & 86.6 & \textbf{68.6} & \textbf{76.5} & 82.9 & \textbf{64.1} & \textbf{72.2} \\
  \hline
  \hline
  \centering \multirow{6}*{VG-200} & SSGRL & 39.7 & 69.9 & 25.9 & 37.8 & 45.3 & 18.3 & 26.1 \\
  \centering ~ & GCN-ML & 39.3 & 64.1 & 28.2 & 38.7 & 44.6 & 18.2 & 25.6 \\
  \centering ~ & KGGR & 41.5 & 64.5 & 30.5 & 41.2 & 54.8 & 25.8 & 33.6 \\
  \centering ~ & Curriculum labeling & 28.4 & 66.4 & 15.4 & 23.6 & 20.4 & 7.6 & 10.9 \\
  \centering ~ & partial-BCE & 39.8 & 69.7 & 24.6 & 36.1 & 44.3 & 18.1 & 25.7 \\
  \centering ~ & Ours & \textbf{45.6} & \textbf{70.1} & \textbf{33.2} & \textbf{45.0} & \textbf{56.8} & \textbf{27.8} & \textbf{37.4} \\
  \hline
  \hline
  \centering \multirow{6}*{Pascal VOC 2007} & SSGRL & 89.5 & 91.2 & \textbf{84.4} & 87.7 & 87.8 & \textbf{81.4} & 84.5 \\
  \centering ~ & GCN-ML & 88.9 & 92.2 & 83.0 & 87.3 & 89.7 & 80.1 & 84.6 \\
  \centering ~ & KGGR & 89.7 & 90.5 & 82.9 & 86.5 & 88.5 & 81.4 & 84.7 \\
  \centering ~ & Curriculum labeling & 84.1 & 92.7 & 78.2 & 83.8 & 79.5 & 71.7 & 75.4 \\
  \centering ~ & partial-BCE & 90.0 & 91.8 & 84.3 & 87.9 & 88.8 & 81.3 & 84.8 \\
  \centering ~ & Ours & \textbf{90.7} & \textbf{93.0} & 83.6 & \textbf{88.4} & \textbf{90.4} & 81.1 & \textbf{85.9} \\
  \hline
  \end{tabular}
  \caption{Average mAP, OP, OR, OF1 and CP, CR, CF1 of the proposed SARB framework and current state-of-the-art competitors for multi-label recognition with partial labels on the MS-COCO, VG-200 and Pascal VOC 2007 datasets. The best results are highlighted in bold.}
  \label{tab:results-detail}
\end{table*}

\begin{table}[!h]
  \centering
  \begin{tabular}{c|ccc}
  \hline
  \centering \diagbox{Methods}{Datasets} & MS-COCO & VG-200 & VOC2007 \\
  \hline
  \hline
  \centering Ours w/ SAM & 77.6 & 45.4 & 90.6 \\
  \centering Ours w/ SD & 77.9 & 45.6 & 90.7 \\
  \centering IP-Mixup & 74.3 & 39.7 & 89.7 \\
  \centering FM-Mixup & 74.1 & 39.6 & 89.6 \\ 
  \hline
  \hline
  \centering SSGRL & 74.1 & 39.7 & 89.5 \\
  \centering Ours ILRB & 77.3 & 44.9 & 90.2 \\
  \centering Ours ILRB fixed $\alpha$ & 76.9 & 44.5  & 89.8 \\
  \centering Ours PLRB & 77.3 & 44.9 & 90.4 \\
  \centering Ours PLRB fixed $\beta$ & 76.9 & 44.6 & 90.2 \\
\hline
  \hline
  \centering Ours & 77.9 & 45.6 & 90.7 \\
  \hline
  \end{tabular}
  \caption{Comparison of average mAP of our framework with SAM module (Ours w/ SAM), our framework with SD module (Ours w/ SD), SSGRL with mixup on image pixel level (IP-Mixup), SSGRL with mixup on feature map level (FM-Mixup), the baseline SSGRL,  our framework merely using ILRB module (Our ILRB), our framework merely using ILRB module with fixed $\alpha$ (Ours ILRB fixed $\alpha$), our framework merely using PLRB module (Ours PLRB), our framework merely using PLRB module with fixed $\beta$ (Ours PLRB fixed $\beta$) and our framework (Ours) on the MS-COCO, VG-200 and Pascal VOC 2007 datasets.}
  \label{tab:ablation-result}
\end{table}

\subsection{Comparison with the state-of-the-art algorithms}

To evaluate the effectiveness of the proposed SARB framework, we compare it with both the conventional MLR and current MLR-PL algorithms: 

\noindent \textit{1) Conventional MLR algorithms}: semantic-specific graph representation learning (SSGRL) \cite{Chen2019SSGRL}, multi-label image recognition graph convolution network (GCN-ML) \cite{Chen2019ML-GCN}, knowledge-guided graph routing (KGGR) \cite{Chen2020KGGR}. Through exploring label dependencies or capturing semantic information, these methods achieve state-of-the-art performance on the traditional MLR task. For fair comparisons, we adapt these methods to address the MLR-PL task by replacing BCE loss with partial BCE loss. 

\noindent \textit{2) Current MLR-PL algorithms}: partial binary cross entropy loss (partial-BCE) \cite{Durand2019CVPR}, Curriculum Labeling \cite{Durand2019CVPR}. It is worth noting that partial-BCE not only is easy to implement but also achieves state-of-the-art performance on the MLR-PL task.

\subsubsection{Performance on MS-COCO} 
We first present the performance comparisons on MS-COCO in Table \ref{tab:results-detail} and Figure \ref{fig:mAP-result-1}. Our SARB framework obtains the overall best performance over current state-of-the-art algorithms. As shown in Table \ref{tab:results-detail}, it achieves the average mAP, OF1, and CF1 of 77.9\%, 76.5\%, and 72.2\%, outperforming the previous best-performing KGGR algorithm by 2.3\%, 2.8\%, and 2.5\%, respectively. As shown in Figure \ref{fig:mAP-result-1}, the SARB framework also achieves better mAP over all known label proportion settings. It is noteworthy that the SARB framework obtains more obvious performance improvement when decreasing the known label proportions. For example, the mAP improvements over the previous best KGGR algorithm are 1.4\% and 4.6\% when using 90\% and 10\% known labels, respectively. These comparisons demonstrate that the SARB framework can be adapted to different proportion settings as it does not depend on pre-trained models.

\subsubsection{Performance on VG-200}
As previously discussed, VG-200 is a more challenging benchmark that covers much more categories. Thus, current works achieve quite poor performance. As shown in Table \ref{tab:results-detail}, the previous best-performing KGGR algorithm obtains the average mAP, OF1, and CF1 of 41.5\%, 41.2\%,and 33.6\%. In this scenario, our SARB framework exhibits much more obvious performance improvement. Its average mAP, OF1, and CF1 are 45.6\%, 45.0\%, and 37.4\%, outperforming the KGGR algorithm by 4.1\%, 3.8\%, and 3.8\%. We also present the mAP comparisons over different known proportion settings in Figure \ref{fig:mAP-result-2}. Compared with current algorithms, we find that our framework achieves the mAP improvement of more than 3.3\% on all known label proportion settings.

\subsubsection{Performance on Pascal VOC 2007}
Pascal VOC 2007 is the most widely used dataset for evaluating multi-label image recognition. Here, we also present the performance comparisons on this dataset in Table \ref{tab:results-detail} and Figure \ref{fig:mAP-result-3}. As this dataset covers merely 20 categories, it is a much simpler dataset and current algorithms can also achieve quite well performance. However, our SARB framework can still achieve consistent improvement. As shown, it improves the average mAP, OF1, and CF1 by 0.7\%, 0.5\%, and 1.1\%. In addition, it exhibits a similar phenomenon that the mAP improvement is more obvious when using the fewer known labels, with 0.4\% and 2.8\% mAP improvement using 90\% and 10\% known label proportions as shown in Figure \ref{fig:mAP-result-3}. 

\section{Ablative Studies}

In this section, we conduct ablative studies to analyze the actual contributions of each module in our SARB framework. 

\subsection{Analysis of the CSRL module}
The CSRL module is used to extract category-specific feature representation and is a basic module of the proposed framework. There are different kinds of algorithms to implement the CSRL module, in which semantic decoupling (SD) \cite{Chen2019SSGRL} and semantic attention mechanism (SAM) \cite{Ye2020ADD-GCN} are two choices that obtain state-of-the-art performance for the traditional MLR task. Here, we conduct an experiment to compare these two algorithms and present the results in Table \ref{tab:ablation-result}. It shows that using the two algorithms obtain comparable performance. More concretely, using SD obtains slightly better performance than using SAM, with an average mAP improvement of 0.3\%, 0.2\%, and 0.1\% on the three datasets. Thus, we use the SD to implement the CSRL module for all other experiments.

Current mixup \cite{Zhang2017Mixup} simply performs position-wise blending to generate new samples to regularize training. In this part, we further conduct two baseline algorithms that perform position-wise blending in image space and feature space (namely IP-Mixup and FM-Mixup) to verify the benefit of learning category-specific feature representation. As shown in Table \ref{tab:ablation-result}, both two baseline algorithms achieve comparable performance with the SSGRL baselines as such simple blending can not provide additional information. Compared with the SARB using CSRL, IP-Mixup suffers from the average mAP degration of 3.6\%, 5.9\%, and 1.0\%, while FM-Mixup suffers from the average mAP degration of 3.8\%, 6.0\%, and 1.1\% on the three datasets, respectively.

\subsection{Contribution of the SARB module}
As we use the SD algorithm to implement the CSRL module and gated neural network for classification, SSGRL \cite{Chen2019SSGRL} is the baseline of the proposed framework. Here, we emphasize the comparisons with SSGRL to demonstrate the effectiveness of SARB. As shown in Table \ref{tab:ablation-result}, SSGRL obtains the average mAPs of 74.1\%, 39.7\%, and 89.5\% on the MS-COCO, VG-200, and Pascal VOC datasets. By integrating the SARB module, it boosts the average mAP to 77.9\%, 45.6\%, and 90.7\% on the three datasets, with the mAP improvement of 3.8\%, 5.9\%, and 1.2\%, respectively.

SARB consists of the instance-level and prototype-level representation blending modules. In the following, we further conduct experiments to analyze these two modules for more in-depth understanding.

\subsection{Analysis of the ILRB module}
To analyze the actual contribution of the ILRB module, we conduct experiments that merely use this module (namely, Ours ILRB) and compare it with the SSGRL baseline on the MS-COCO, VG-200, Pascal VOC 2007 datasets. As shown in Table \ref{tab:ablation-result}, it obtains an average mAP of 77.3\%, 44.9\%, 90.2\% on MS-COCO, VG-200, Pascal VOC 2007, with the mAP improvement of 3.2\%, 5.2\%, and 0.7\%, respectively. 

ILRB contains an crucial parameter $\alpha$ that controls the ratio of instance-level mix-up. However, it is impractical and exhausting to find a best value for different datasets and different settings. In this work, we set $\alpha$ as a learnable parameter to adaptively learn the best value via standard back-propagation. To verify its contribution, we conduct an experiment to compare with the baseline using a fixed $\alpha$ of 0.5. As shown in Table \ref{tab:ablation-result}, using a fixed value of 0.5 decreases the average mAPs from 77.3\%, 44.9\%, and 90.2\% to 76.9\%, 44.5\%, and 89.8\%, respectively.

\begin{figure}[!t] 
  \centering    
  \subfigure {
  \includegraphics[width=0.47\linewidth]{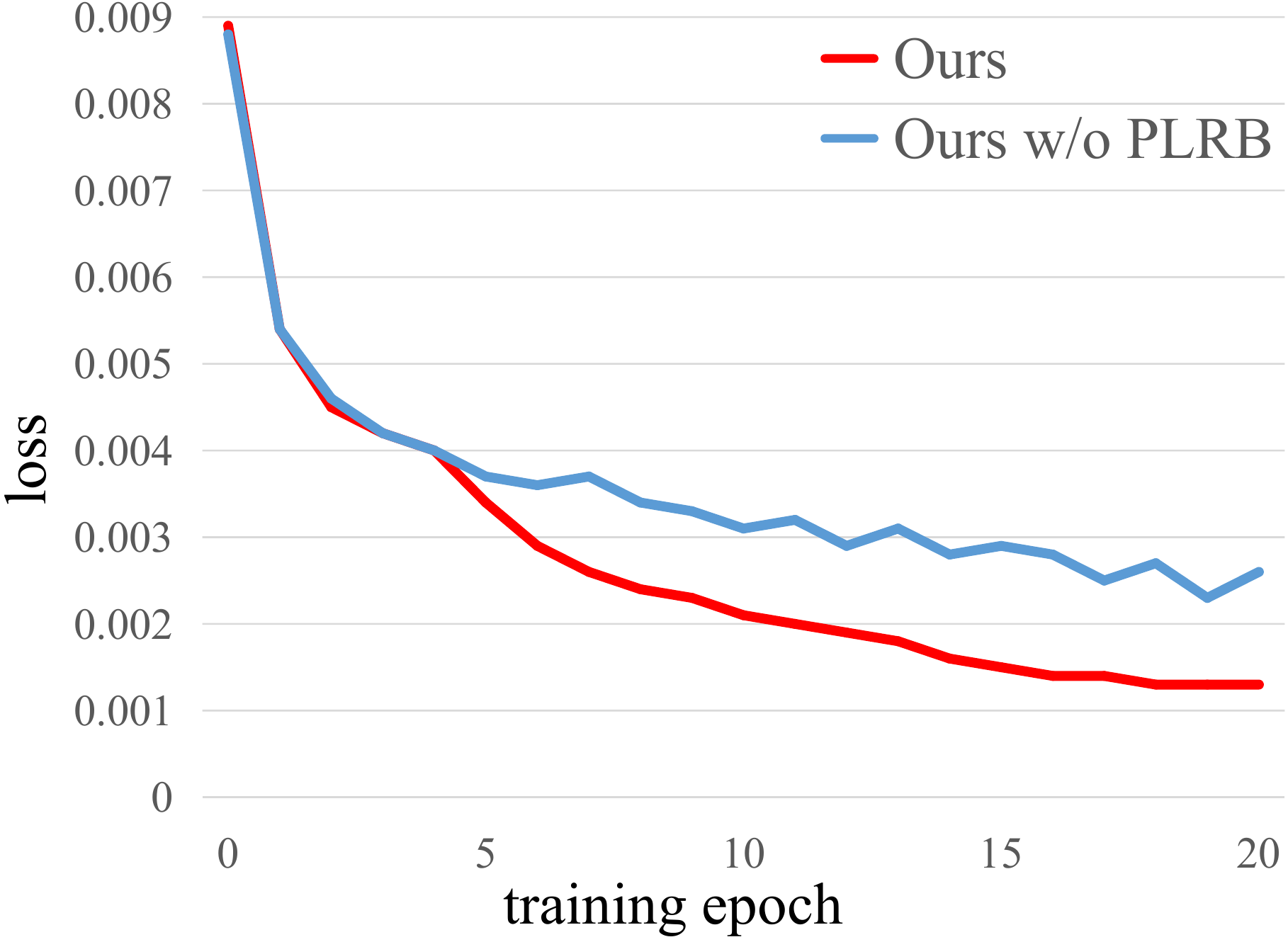}  
  }~     
  \subfigure { 
  \includegraphics[width=0.47\linewidth]{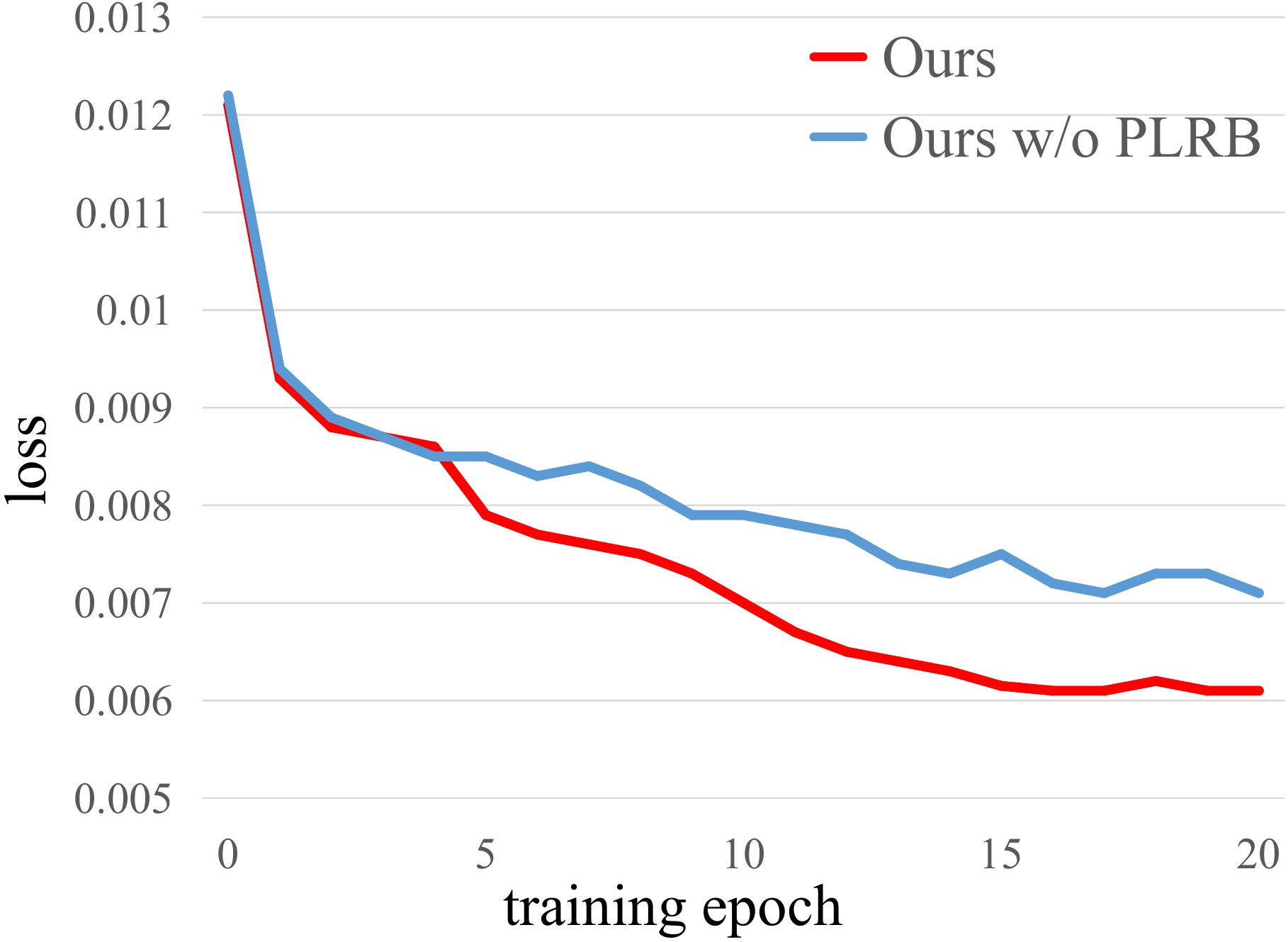}     
  }~
  \caption{Analysis of the effect on PLRB. These experiments are conducted on MS-COCO (left) and VG-200 (right).} 
  \label{fig:loss-result}     
\end{figure}

\subsection{Analysis of the PLRB module}
Similarly, PLRB is another module that plays a key role, and in this part, we also analyze its effectiveness by comparing the performance with and without it. As shown in Table \ref{tab:ablation-result}. Adding the PLRB module to the baseline SSGRL leads to 3.2\%, 5.2\%, and 0.9\% mAP improvement on the MS-COCO, VG-200, and Pascal VOC 2007 datasets. As previously suggested, the PLRB module can help to generate stable blended representations to complement unknown labels, which leads to more stable training. To validate this point, we further visualize the loss of the training process in Figure \ref{fig:loss-result}. It can be observed that the loss is choppy without the PLRB module, and adding this module can stabilize the training process.

The parameter $\beta$ is a learnable parameter that is adaptively learned for different datasets and settings. Here, we also conduct experiments to compare with the setting that fixes $\beta$ to 0.5 on the MS-COCO, VG-200, Pascal VOC 2007 datasets. As presented in Table \ref{tab:ablation-result}, it obtains an average mAP of 76.9\%, 44.6\%, 90.2\% on these three datasets, with the slight degeneration of 0.4\%, 0.3\% and 0.2\%. 

\section{Conclusion}
In this work, we present a new perspective to complement the unknown labels by blending category-specific feature representation to address the MLR-PL task. It does not depend on sufficient annotations and thus can obtain superior performance on all known label proportion settings. Specifically, it consists of an ILRB module that blends instance-level representation of known labels to complement the representation of corresponding unknown labels and a PLRB module that leans and blends prototype-level representations to complement the representation of corresponding unknown labels. It can simultaneously generate diverse and stable blended representations to complement the unknown labels and thus facilitate the MLR-PL task. Extensive experiments on the MS-COCO, VG-200, and Pascal VOC demonstrate its superiority over current algorithms.

\section{Acknowledgements}
This work was supported by National Natural Science Foundation of China (No. 61876045, 61836012 and 62002069), the Natural Science Foundation of Guangdong Province (No. 2017A030312006) and Guangdong Provincial Basic Research Program (No. 102020369).

{\small
\bibliography{aaai22}
}

\clearpage

\section{Supplementary Material}

Due to the limit of the page range, we present some definition, detailed results in this supplementary material.

\subsection{Evaluation Metric}
In the original paper, we present the mAP and average of the overall precision, recall, F1-measure (OP, OR, OF1) and per-class precision, recall, F1-measure (CP, CR, CF1) for evaluation. There, we present the definitions of the OP, OR, OF1 and CP, CR, CF1 in detals. Formally, these metrics can be computed by
\begin{gather}
 OP=\frac{\sum_{i}{N^c_i}}{\sum_{i}{N^p_i}}, CP=\frac{1}{C} \sum_{i}{ \frac{N^c_i}{N^p_i} } \\
 OR=\frac{\sum_{i}{N^c_i}}{\sum_{i}{N^g_i}}, CR=\frac{1}{C} \sum_{i}{ \frac{N^c_i}{N^g_i} } \\
 OF1=\frac{2 \times OP \times OR }{OP+OR},  CF1=\frac{2 \times CP \times CR }{CP+CR}
\end{gather}
where $N^c_i$ is the number of images that are correctly predicted for the $i$-th label, $N^p_i$ is the number of predicted images for the $i$-th label, $N^g_i$ is the number of ground truth images for the $i$-th label. We also average the OP, OR, OF1, CP, CR, CF1 over all known label proportions. 

\subsection{Performance on the mAP metric}
In the original paper, we have present the mAP comparisions with SSGRL \cite{Chen2019SSGRL}, GCN-ML \cite{Chen2019ML-GCN}, KGGR \cite{Chen2020KGGR}, partial-BCE \cite{Durand2019CVPR}, and Curriculum Labeling \cite{Durand2019CVPR} on all known label proportion settings. However, we merely present mAP-proportion curve for direct comparison. For a more detailed comparisons, we also present the mAP values of all the algorithms over all settings on the MS-COCO \cite{Lin2014COCO}, VG-200 \cite{Krishna2016VG}, and Pascal VOC 2007 \cite{Everingham2010Pascal} datasets in Table \ref{tab:mAP-results}.

\subsection{Performance on the OF1 and CF1 metric}
\begin{figure*}[!t]
\centering
\subfigure{
\label{fig:coco-of1}
\includegraphics[width=0.31\linewidth]{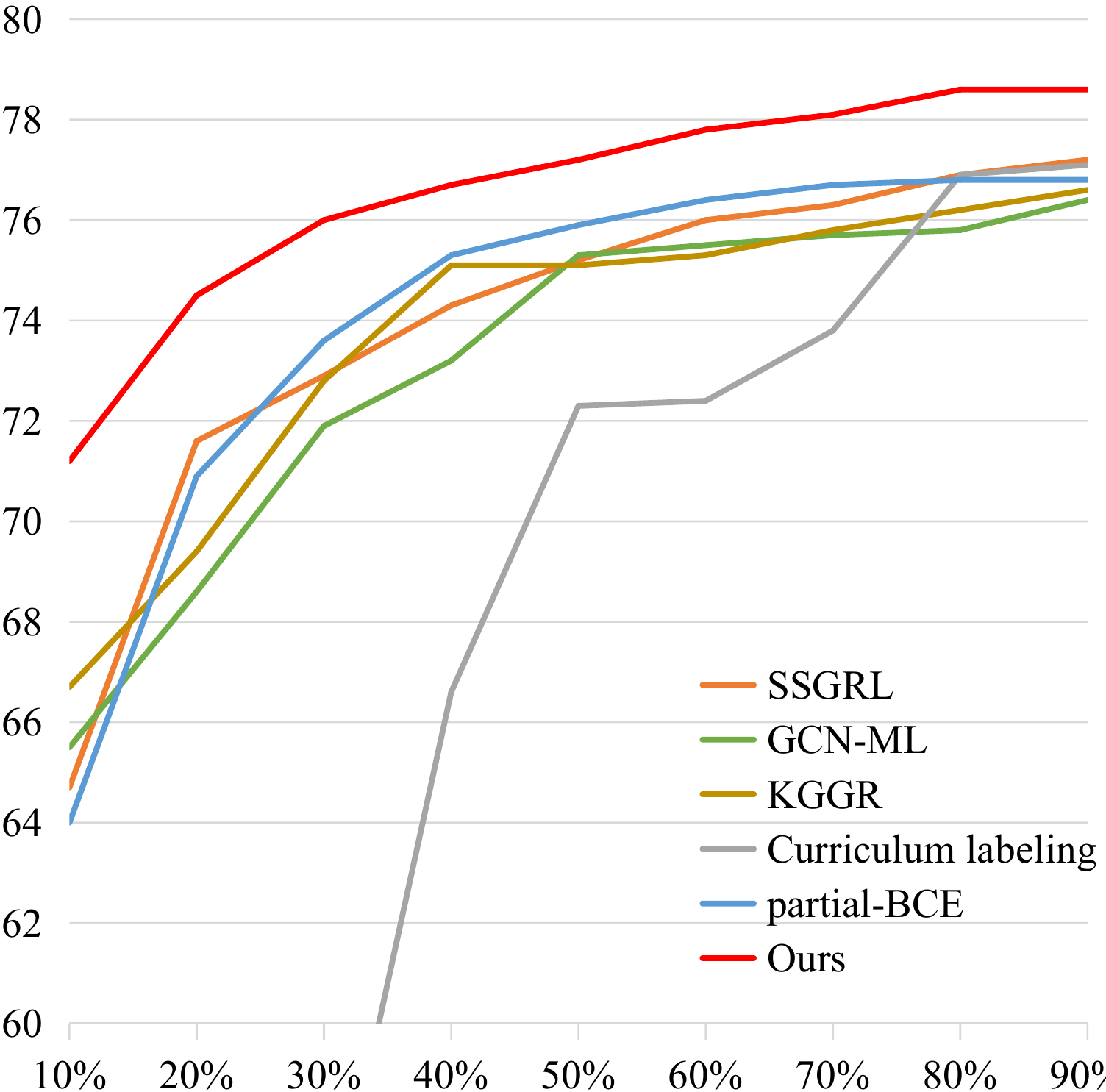}}
\subfigure{
\label{fig:vg-of1}
\includegraphics[width=0.31\linewidth]{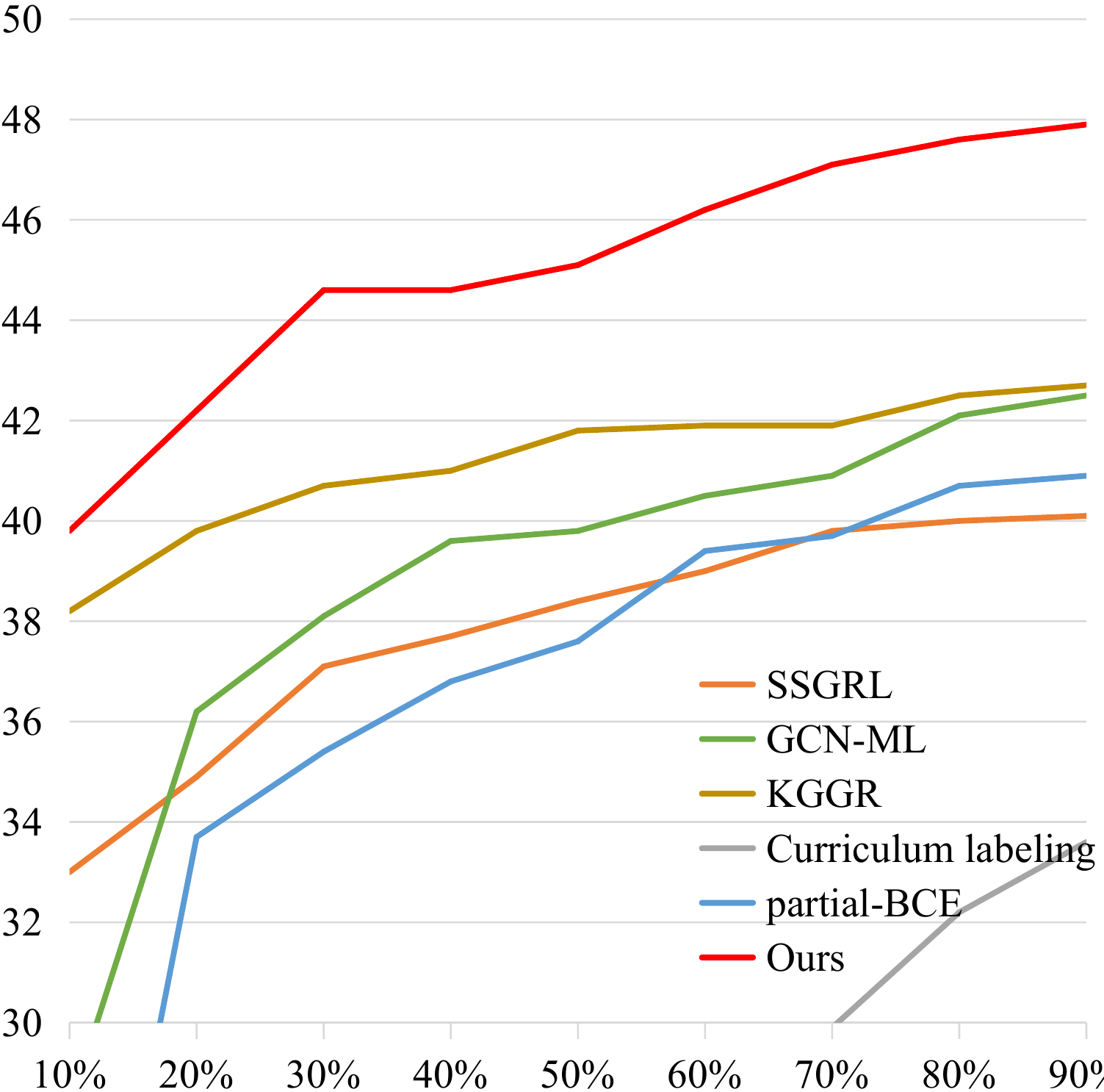}}
\subfigure{
\label{fig:voc-of1}
\includegraphics[width=0.31\linewidth]{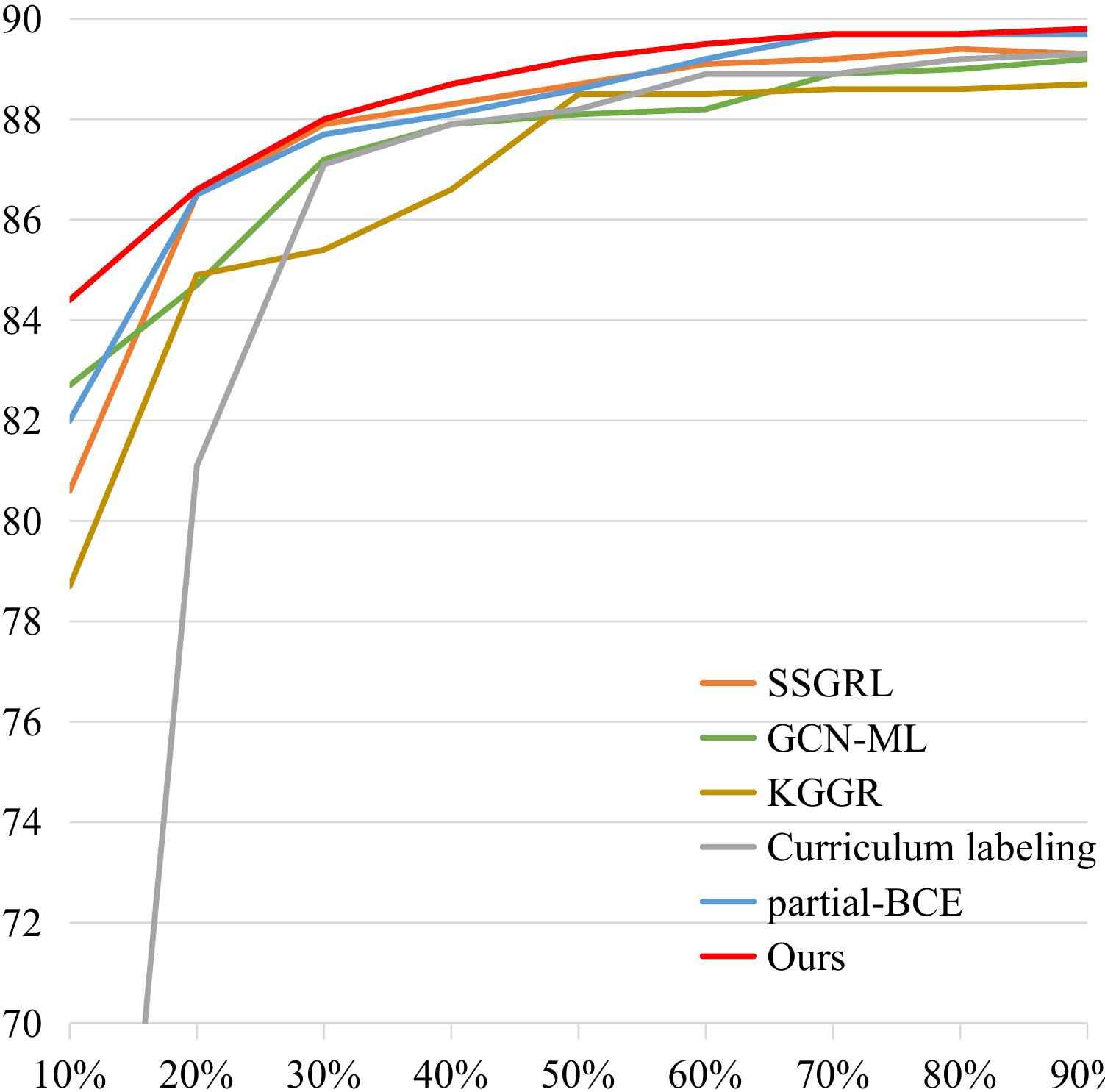}}
\subfigure{
\label{fig:coco-cf1} 
\includegraphics[width=0.31\linewidth]{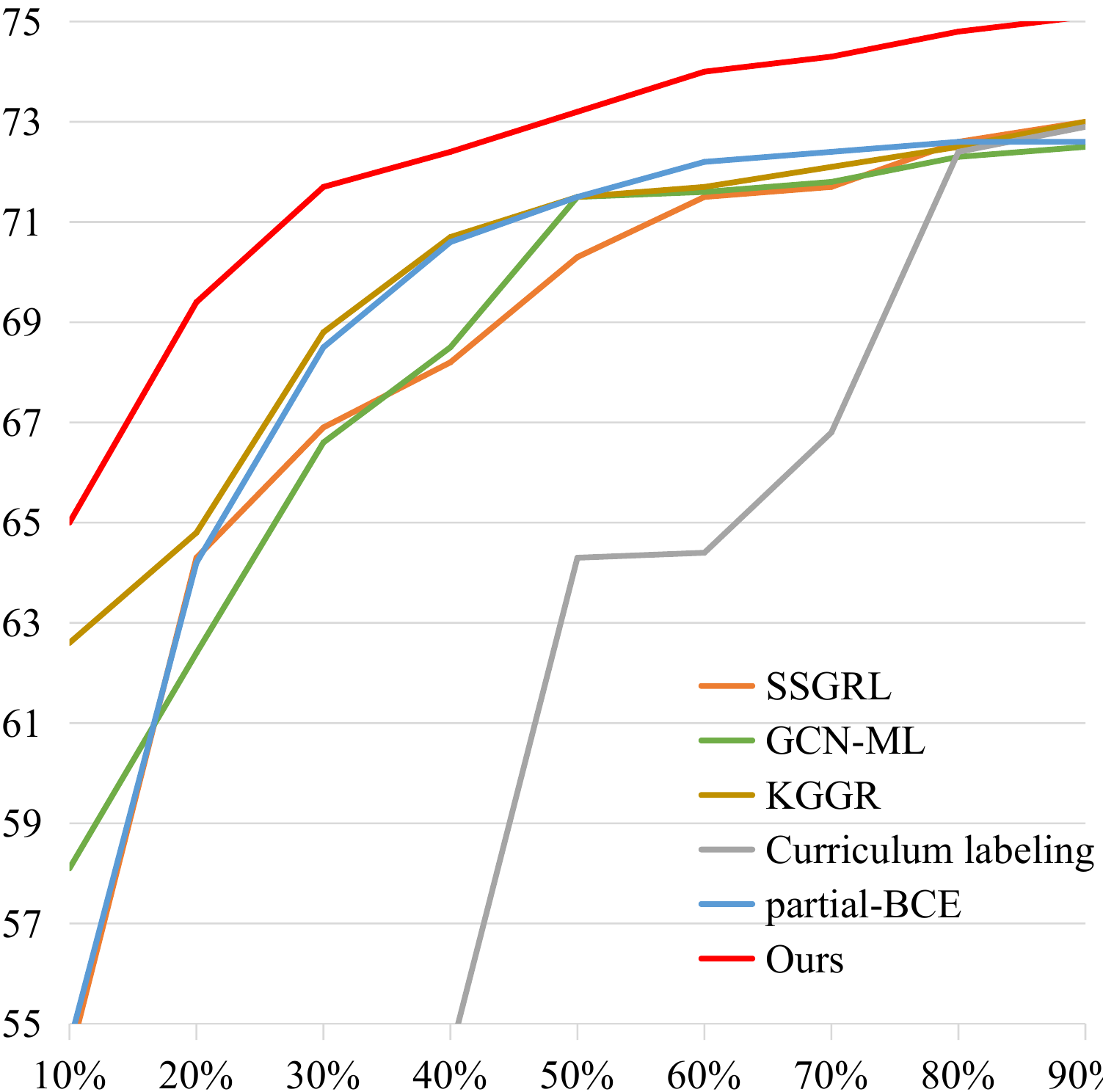}}
\subfigure{
\label{fig:vg-cf1}
\includegraphics[width=0.31\linewidth]{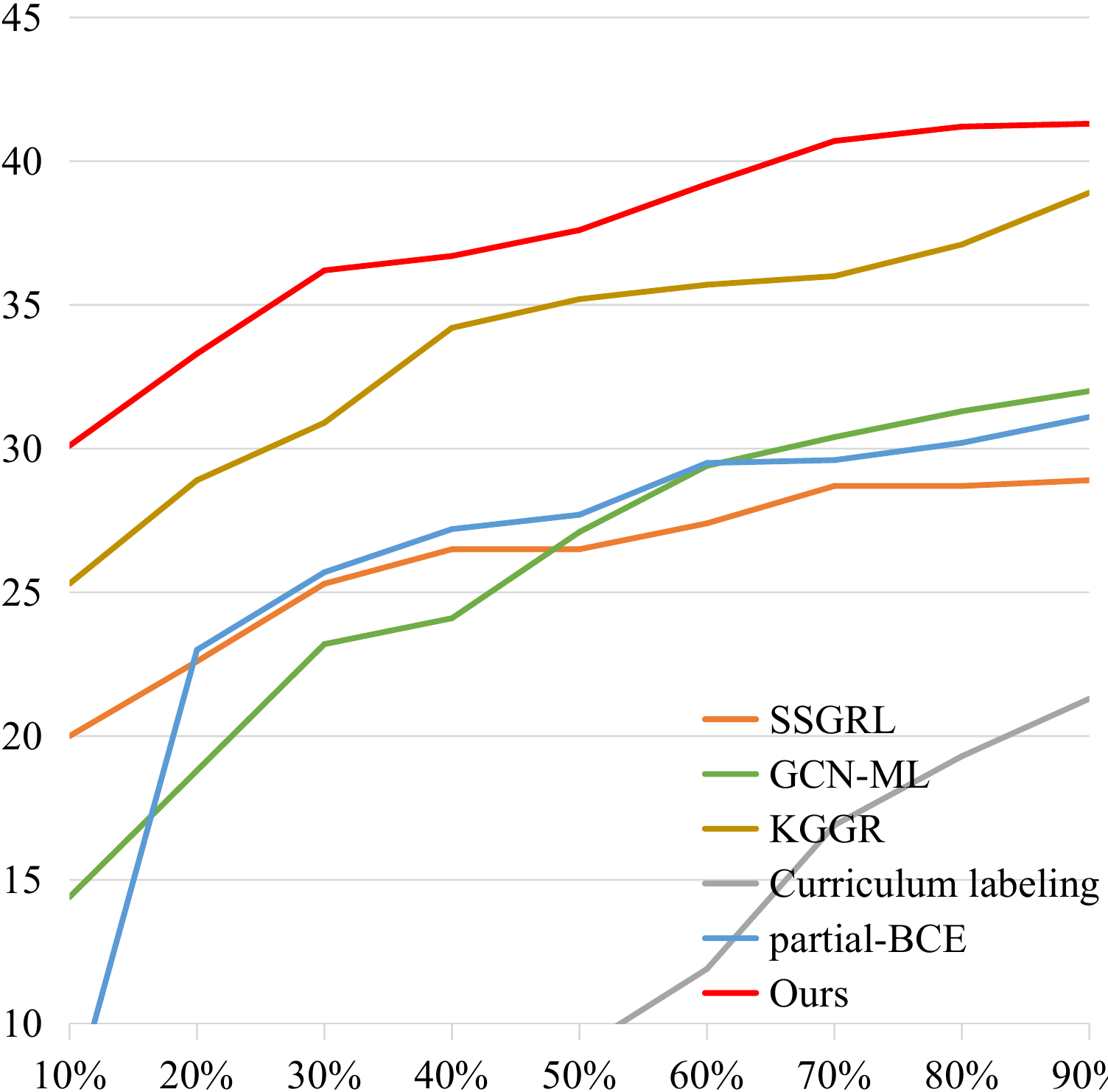}}
\subfigure{
\label{fig:voc-cf1} 
\includegraphics[width=0.31\linewidth]{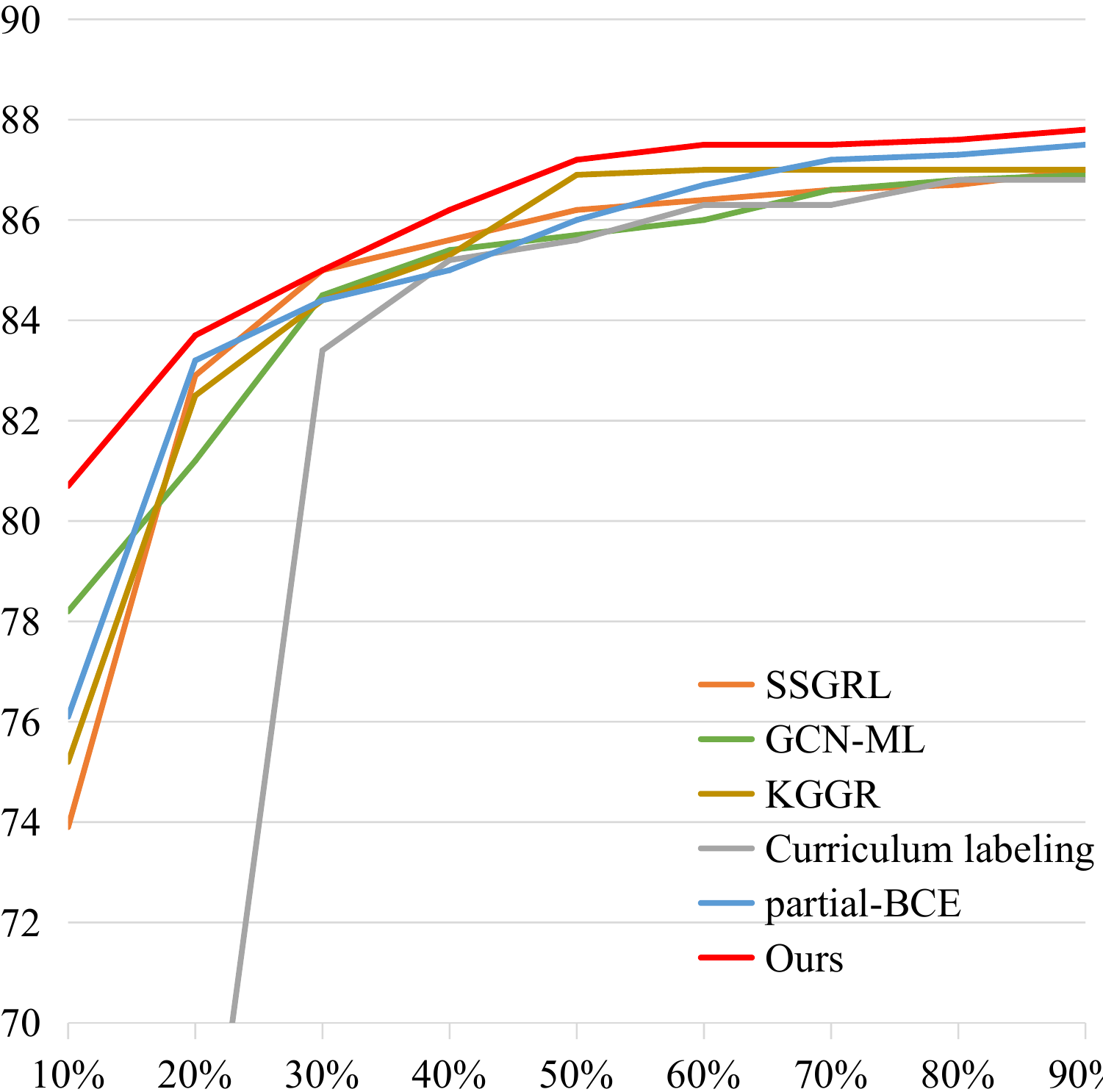}}
\vspace{0pt}
\caption{The OF1 (first row) and CF1 (second row) over different known label proportions of 10\%, 20\%, ..., 90\% of the proposed SARB framework and current state-of-the-art competitors on the MS-COCO (left), VG-200 (middle), and Pascal VOC 2007 (right) datasets.}
\label{fig:of1-cf1}
\end{figure*}

In the original paper, we have also presented the average OP, OR, OF1 and CP, CR CF1 for a more comprehensive comparison, but we do not provide these metrics on different known label proportion settings due to the page limit. Here, we first present the detailed OF1 and CF1 on all known label proportion settings in Figure \ref{fig:of1-cf1} and Table \ref{tab:OF1-CF1-results}. As shown, our SARB framework can also achieve consistently better OF1 and CF1 for all known label proportion settings. For example on the MS-COCO dataset, our SARB methods achieve the average OF1 and CF1 of 76.5\% and 72.2\%. It obtains up to 5.1\% OF1 improvement and up to 4.6\% CF1 improvement compared with the previous best-performing KGGR algorithm. On the more challenging VG-200 dataset, it obtain the even more obvious OF1 improvement ranging from 2.4\% to 6.8\% and CF1 improvement ranging from 2.4\% to 5.3\% compared with the current state-of-the-art KGGR algorithm on different known label proportion settings. On the Pascal VOC 2007 dataset, our SARB framework also obtains evident improvement, up to 1.7\% and 2.5\% OF1 and CF1 improvement with 10\% known labels. 

\subsection{Analysis of parameters $\alpha$ and $\beta$}
$\alpha$ and $\beta$ are two crucial parameters in the ILRB and PLRB modules, respectively. In the original paper, we propose to learn these two parameters thus as to avoid fine-tuning them for different datasets and different known label proportion settings. We have also compared the performance of learning these two parameters and directly setting them to 0.5. Here, we further conduct experiment that sets the parameters $\alpha$ and $\beta$ to different values (i.e., 0.1, 0.3, 0.5, 0.7, and 0.9). As shown in Table \ref{tab:ablation-result}, we find setting $\alpha$ and $\beta$ to 0.5 obtains the best performance. Thus, we simply compare the learnable setting with setting them to 0.5 in the original paper.

\begin{table*}[!t]
  \centering
  \small
  \begin{tabular}{c|c|ccccccccc|c}
  \hline
  \centering Datasets & Methods & 10\% & 20\% & 30\% & 40\% & 50\% & 60\% & 70\% & 80\% & 90\% & Ave. mAP \\
  \hline
  \hline
  \centering \multirow{6}*{MS-COCO} & SSGRL & 62.5 & 70.5 & 73.2 & 74.5 & 76.3 & 76.5 & 77.1 & 77.9 & 78.4 & 74.1 \\
  \centering ~ & GCN-ML & 63.8 & 70.9 & 72.8 & 74.0 & 76.7 & 77.1 & 77.3 & 78.3 & 78.6 & 74.4 \\
  \centering ~ & KGGR & 66.6 & 71.4 & 73.8 & 76.7 & 77.5 & 77.9 & 78.4 & 78.7 & 79.1 & 75.6 \\
  \centering ~ & Curriculum labeling & 26.7 & 31.8 & 51.5 & 65.4 & 70.0 & 71.9 & 74.0 & 77.4 & 78.0 & 60.7 \\
  \centering ~ & partial-BCE & 61.6 & 70.5 & 74.1 & 76.3 & 77.2 & 77.7 & 78.2 & 78.4 & 78.5 & 74.7 \\
  \centering ~ & Ours & \textbf{71.2} & \textbf{75.0} & \textbf{77.1} & \textbf{78.3} & \textbf{78.9} & \textbf{79.6} & \textbf{79.8} & \textbf{80.5} & \textbf{80.5} & \textbf{77.9} \\
  \hline
  \hline
  \centering \multirow{6}*{VG-200} & SSGRL & 34.6 & 37.3 & 39.2 & 40.1 & 40.4 & 41.0 & 41.3 & 41.6 & 42.1 & 39.7 \\
  \centering ~ & GCN-ML & 32.0 & 37.8 & 38.8 & 39.1 & 39.6 & 40.0 & 41.9 & 42.3 & 42.5 & 39.3 \\
  \centering ~ & KGGR & 36.0 & 40.0 & 41.2 & 41.5 & 42.0 & 42.5 & 43.3 & 43.6 & 43.8 & 41.5 \\
  \centering ~ & Curriculum labeling & 12.1 & 19.1 & 25.1 & 26.7 & 30.0 & 31.7 & 35.3 & 36.8 & 38.5 & 28.4 \\
  \centering ~ & partial-BCE & 27.4 & 38.1 & 40.2 & 40.9 & 41.5 & 42.1 & 42.4 & 42.7 & 42.7 & 39.8 \\
  \centering ~ & Ours & \textbf{40.6} & \textbf{43.5} & \textbf{44.5} & \textbf{45.3} & \textbf{46.0} & \textbf{47.1} & \textbf{47.2} & \textbf{47.8} & \textbf{48.1} & \textbf{45.6} \\
  \hline
  \hline
  \centering \multirow{6}*{Pascal VOC 2007} & SSGRL & 77.7 & 87.6 & 89.9 & 90.7 & 91.4 & 91.8 & 91.9 & 92.2 & 92.2 & 89.5 \\
  \centering ~ & GCN-ML & 74.5 & 87.4 & 89.7 & 90.7 & 91.0 & 91.3 & 91.5 & 91.8 & 92.0 & 88.9 \\
  \centering ~ & KGGR & 81.3 & 88.1 & 89.9 & 90.4 & 91.2 & 91.3 & 91.5 & 91.6 & 91.8 & 89.7 \\
  \centering ~ & Curriculum labeling & 44.7 & 76.8 & 88.6 & 90.2 & 90.7 & 91.1 & 91.6 & 91.7 & 91.9 & 84.1 \\
  \centering ~ & partial-BCE & 80.7 & 88.4 & 89.9 & 90.7 & 91.2 & 91.8 & 92.3 & 92.4 & 92.5 & 90.0 \\
  \centering ~ & Ours & \textbf{83.5} & \textbf{88.6} & \textbf{90.7} & \textbf{91.4} & \textbf{91.9} & \textbf{92.2} & \textbf{92.6} & \textbf{92.8} & \textbf{92.9} & \textbf{90.7} \\
  \hline
  \end{tabular}
  \vspace{0pt}
  \caption{The mAP of the proposed SARB framework and current state-of-the-art competitors for MLR-PL on the MS-COCO, VG-200 and Pascal VOC 2007 datasets. The experiments are conducted with 10\%-90\% proportions of known labels. The best results are highlighted in bold.}
  \label{tab:mAP-results}
\end{table*}

\begin{table*}[!t]
  \centering
  \small
  \begin{tabular}{c|c|c|ccccccccc|c}
  \hline
  \centering Datasets & Metrics & Methods & 10\% & 20\% & 30\% & 40\% & 50\% & 60\% & 70\% & 80\% & 90\% & Average \\
  \hline
  \hline
  \centering \multirow{12}*{MS-COCO} & \multirow{6}*{OF1} & SSGRL & 64.7 & 71.6 & 72.9 & 74.3 & 75.2 & 76.0 & 76.3 & 76.9 & 77.2 & 73.9 \\
  \centering ~ & ~ & GCN-ML & 65.5 & 68.6 & 71.9 & 73.2 & 75.3 & 75.5 & 75.7 & 75.8 & 76.4 & 73.1 \\
  \centering ~ & ~ & KGGR & 66.7 & 69.4 & 72.8 & 75.1 & 75.1 & 75.3 & 75.8 & 76.2 & 76.6 & 73.7 \\
  \centering ~ & ~ & Curriculum labeling & 29.0 & 33.8 & 55.0 & 66.6 & 72.3 & 72.4 & 73.8 & 76.9 & 77.1 & 61.9 \\
  \centering ~ & ~ & partial-BCE & 64.0 & 70.9 & 73.6 & 75.3 & 75.9 & 76.4 & 76.7 & 76.8 & 76.8 & 74.0 \\
  \centering ~ & ~ & Ours & \textbf{71.2} & \textbf{74.5} & \textbf{76.0} & \textbf{76.7} & \textbf{77.2} & \textbf{77.8} & \textbf{78.1} & \textbf{78.6} & \textbf{78.6} & \textbf{76.5} \\
  \cline{2-13}
  \centering ~ & \multirow{6}*{CF1} & SSGRL & 54.3 & 64.3 & 66.9 & 68.2 & 70.3 & 71.5 & 71.7 & 72.6 & 73.0 & 68.1 \\
  \centering ~ & ~ & GCN-ML & 58.1 & 62.4 & 66.6 & 68.5 & 71.5 & 71.6 & 71.8 & 72.3 & 72.5 & 68.4 \\
  \centering ~ & ~ & KGGR & 62.6 & 64.8 & 68.8 & 70.7 & 71.5 & 71.7 & 72.1 & 72.5 & 73.0 & 69.7 \\
  \centering ~ & ~ & Curriculum labeling & 3.7 & 3.7 & 32.4 & 54.4 & 64.3 & 64.4 & 66.8 & 72.4 & 72.9 & 48.3 \\
  \centering ~ & ~ & partial-BCE & 54.6 & 64.2 & 68.5 & 70.6 & 71.5 & 72.2 & 72.4 & 72.6 & 72.6 & 68.8 \\
  \centering ~ & ~ & Ours & \textbf{65.0} & \textbf{69.4} & \textbf{71.7} & \textbf{72.4} & \textbf{73.2} & \textbf{74.0} & \textbf{74.3} & \textbf{74.8} & \textbf{75.1} & \textbf{72.2} \\
  \hline
  \hline
  \centering \multirow{12}*{VG-200} & \multirow{6}*{OF1} & SSGRL & 33.0 & 34.9 & 37.1 & 37.7 & 38.4 & 39.0 & 39.8 & 40.0 & 40.1 & 37.8 \\
  \centering ~ & ~ & GCN-ML & 28.3 & 36.2 & 38.1 & 39.6 & 39.8 & 40.5 & 40.9 & 42.1 & 42.5 & 38.7 \\
  \centering ~ & ~ & KGGR & 38.2 & 39.8 & 40.7 & 41.0 & 41.8 & 41.9 & 41.9 & 42.5 & 42.7 & 41.2 \\
  \centering ~ & ~ & Curriculum labeling & 9.2 & 14.8 & 20.4 & 22.1 & 23.4 & 26.6 & 29.9 & 32.2 & 33.6 & 23.6 \\
  \centering ~ & ~ & partial-BCE & 20.8 & 33.7 & 35.4 & 36.8 & 37.6 & 39.4 & 39.7 & 40.7 & 40.9 & 36.1 \\
  \centering ~ & ~ & Ours & \textbf{39.8} & \textbf{42.2} & \textbf{44.6} & \textbf{44.6} & \textbf{45.1} & \textbf{46.2} & \textbf{47.1} & \textbf{47.6} & \textbf{47.9} & \textbf{45.0} \\
  \cline{2-13}
  \centering ~ & \multirow{6}*{CF1} & SSGRL & 20.0 & 22.6 & 25.3 & 26.5 & 26.5 & 27.4 & 28.7 & 28.7 & 28.9 & 26.1 \\
  \centering ~ & ~ & GCN-ML & 14.4 & 18.8 & 23.2 & 24.1 & 27.1 & 29.4 & 30.4 & 31.3 & 32.0 & 25.6 \\
  \centering ~ & ~ & KGGR & 25.3 & 28.9 & 30.9 & 34.2 & 35.2 & 35.7 & 36.0 & 37.1 & 38.9 & 33.6 \\
  \centering ~ & ~ & Curriculum labeling & 2.1 & 3.3 & 6.8 & 7.3 & 9.1 & 11.9 & 16.9 & 19.3 & 21.3 & 10.9 \\
  \centering ~ & ~ & partial-BCE & 6.9 & 23.0 & 25.7 & 27.2 & 27.7 & 29.5 & 29.6 & 30.2 & 31.1 & 25.7 \\
  \centering ~ & ~ & Ours & \textbf{30.1} & \textbf{33.3} & \textbf{36.2} & \textbf{36.7} & \textbf{37.6} & \textbf{39.2} & \textbf{40.7} & \textbf{41.2} & \textbf{41.3} & \textbf{37.4} \\
  \hline
  \hline
  \centering \multirow{12}*{Pascal VOC 2007} & \multirow{6}*{OF1} & SSGRL & 80.6 & 86.5 & 87.9 & 88.3 & 88.7 & 89.1 & 89.2 & 89.4 & 89.3 & 87.7 \\
  \centering ~ & ~ & GCN-ML & 82.7 & 84.7 & 87.2 & 87.9 & 88.1 & 88.2 & 88.9 & 89.0 & 89.2 & 87.3 \\
  \centering ~ & ~ & KGGR & 78.7 & 84.9 & 85.4 & 86.6 & 88.5 & 88.5 & 88.6 & 88.6 & 88.7 & 86.5 \\
  \centering ~ & ~ & Curriculum labeling & 53.9 & 81.1 & 87.1 & 87.9 & 88.2 & 88.9 & 88.9 & 89.2 & 89.3 & 83.8 \\
  \centering ~ & ~ & partial-BCE & 82.0 & 86.5 & 87.7 & 88.1 & 88.6 & 89.2 & 89.7 & 89.7 & 89.7 & 87.9 \\
  \centering ~ & ~ & Ours & \textbf{84.4} & \textbf{86.6} & \textbf{88.0} & \textbf{88.7} & \textbf{89.2} & \textbf{89.5} & \textbf{89.7} & \textbf{89.7} & \textbf{89.8} & \textbf{88.4} \\
  \cline{2-13}
  \centering ~ & \multirow{6}*{CF1} & SSGRL & 73.9 & 82.9 & 85.0 & 85.6 & 86.2 & 86.4 & 86.6 & 86.7 & 87.0 & 84.5 \\
  \centering ~ & ~ & GCN-ML & 78.2 & 81.2 & 84.5 & 85.4 & 85.7 & 86.0 & 86.6 & 86.8 & 86.9 & 84.6 \\
  \centering ~ & ~ & KGGR & 75.2 & 82.5 & 84.4 & 85.3 & 86.9 & 87.0 & 87.0 & 87.0 & 87.0 & 84.7 \\
  \centering ~ & ~ & Curriculum labeling & 13.5 & 64.5 & 83.4 & 85.2 & 85.6 & 86.3 & 86.3 & 86.8 & 86.8 & 75.4 \\
  \centering ~ & ~ & partial-BCE & 76.1 & 83.2 & 84.4 & 85.0 & 86.0 & 86.7 & 87.2 & 87.3 & 87.5 & 84.8 \\
  \centering ~ & ~ & Ours & \textbf{80.7} & \textbf{83.7}  & \textbf{85.0} & \textbf{86.2} & \textbf{87.2} & \textbf{87.5} & \textbf{87.5} & \textbf{87.6} & \textbf{87.8} & \textbf{85.9} \\
  \hline
  \end{tabular}
  \vspace{0pt}
  \caption{The OF1 and CF1 of the proposed SARB framework and current state-of-the-art competitors for MLR-PL on the MS-COCO, VG-200 and Pascal VOC 2007 datasets. The experiments are conducted with 10\%-90\% proportions of known labels. The best results are highlighted in bold.}
  \label{tab:OF1-CF1-results}
\end{table*}

\begin{table*}[!h]
  \centering
  \begin{tabular}{c|c|ccccc}
  \hline
  \centering Datasets & Methods & 0.1 & 0.3 & 0.5 & 0.7 & 0.9 \\
  \hline
  \hline
  \centering \multirow{2}*{MS-COCO} & Ours ILRB fixed $\alpha$ & 75.9 & 76.3 & \textbf{76.9} & 76.5 & 76.3 \\
  \centering ~ & Ours PLRB fixed $\beta$ & 75.7 & 76.6 & \textbf{76.9} & 76.7 & 76.4 \\
  \hline
  \hline
  \centering \multirow{2}*{VG-200} & Ours ILRB fixed $\alpha$ & 43.7 & 44.1 & \textbf{44.5} & 44.3 & 44.2 \\
  \centering ~ & Ours PLRB fixed $\beta$ & 43.8 & 44.0 & \textbf{44.6} & 44.3 & 44.1 \\
  \hline
  \hline
  \centering \multirow{2}*{Pascal VOC 2007} & Ours ILRB fixed $\alpha$ & 88.9 & 89.3 & \textbf{89.8} & 89.5 & 89.2 \\
  \centering ~ & Ours PLRB fixed $\beta$ & 89.1 & 89.4 & \textbf{90.2} & 89.7 & 89.3 \\
  \hline
  \end{tabular}
  \caption{The performance of “Ours ILRB fixed $\alpha$” and “Ours PLRB fixed $\beta$” with different initial values of $\alpha$ and $\beta$.}
  \label{tab:ablation-result}
\end{table*}

\end{document}